%% file: main.tex
\pdfoutput=1%
\documentclass{article}

\usepackage{microtype}
\usepackage{graphicx}
\usepackage{subcaption}
\usepackage{booktabs}
\usepackage{placeins}%
\usepackage{hyperref}

\usepackage[preprint]{icml2026}
\hypersetup{pdftitle={Latent Fact Composition under Next-Token Training}, pdfauthor={Tianyi Zhang, Feiying Chen},
  pdfsubject={}}%
\usepackage{amsmath}
\usepackage{amssymb}
\usepackage{mathtools}
\usepackage{amsthm}
\usepackage[capitalize,noabbrev]{cleveref}
\usepackage{tikz}
\usetikzlibrary{arrows.meta,calc}

\input{macros}

\icmltitlerunning{Latent Fact Composition under Next-Token Training}

\begin{document}

\twocolumn[
  \icmltitle{Latent Fact Composition under Next-Token Training}

  \icmlsetsymbol{equal}{*}
  \begin{icmlauthorlist}
    \icmlauthor{Tianyi Zhang}{ucsd}
    \icmlauthor{Feiying Chen}{thu}
  \end{icmlauthorlist}
  \icmlaffiliation{ucsd}{University of California San Diego}
  \icmlaffiliation{thu}{Tsinghua University}
  \icmlcorrespondingauthor{Tianyi Zhang}{tiz054@ucsd.edu}

  \icmlkeywords{looped transformers, latent multi-hop reasoning, compositional generalization, next-token prediction, label noise}
  \vskip 0.3in
]
\printAffiliationsAndNotice{}

\begin{abstract}
\input{sections/abstract}
\end{abstract}

\input{sections/intro}
\input{sections/setting}
\input{sections/architecture}
\input{sections/hypothesis}
\input{sections/decomposition}
\input{sections/loss-results}
\input{sections/related}
\input{sections/limitations}
\input{sections/conclusion}

\bibliography{refs}
\bibliographystyle{icml2026}

\newpage
\appendix
\onecolumn
\input{sections/appendix}

\end{document}

%% file: macros.tex
\newcommand{\archfam}{Baton}%
\newcommand{\mbase}{Baton-H}%
\newcommand{\mq}{Baton-Q}%
\newcommand{\mqal}{Baton-Q\textsubscript{0}}%
\newcommand{\bloop}{\textsc{Loop}}%
\newcommand{\btf}{\textsc{TF8}}%
\newcommand{\bdisco}{DiscoLoop}%

\newcommand{\GA}{\ensuremath{G_A}}%
\newcommand{\GB}{\ensuremath{G_B}}%
\newcommand{\qmodel}{\ensuremath{q_\theta}}%
\newcommand{\ptrain}{\ensuremath{p_{\mathrm{train}}}}%
\newcommand{\pref}{\ensuremath{p_{\mathrm{ref}}}}%
\newcommand{\Htrain}{\ensuremath{H_{\mathrm{train}}}}%
\newcommand{\Res}{\ensuremath{R}}%
\newcommand{\Cgen}{\ensuremath{C_{\mathrm{gen}}}}%
\newcommand{\Cmem}{\ensuremath{C_{\mathrm{mem}}}}%
\newcommand{\KL}{\ensuremath{\mathrm{KL}}}
\newcommand{\E}{\ensuremath{\mathbb{E}}}
\newcommand{\tok}[1]{\texttt{#1}}%

\usepackage{xcolor}
\newif\ifdraft
\draftfalse
\ifdraft
  \newcommand{\todo}[1]{{\color{red!70!black}\textbf{[TODO: #1]}}}
  \newcommand{\zty}[1]{{\color{blue!70!black}[zty: #1]}}
  \newcommand{\claude}[1]{{\color{green!45!black}[Claude: #1]}}
\else
  \newcommand{\todo}[1]{}
  \newcommand{\zty}[1]{}
  \newcommand{\claude}[1]{}
\fi

\newcommand{\shade}[1]{{\setlength{\fboxsep}{1.5pt}\colorbox{gray!20}{#1}}}%
\definecolor{vMtwo}{HTML}{2A78D6}
\definecolor{vMeight}{HTML}{EB6834}
\definecolor{vMeightal}{HTML}{1BAF7A}

%% file: sections/abstract.tex
Controlled studies of latent two-hop reasoning train transformers to predict only the answer, whereas language models are trained to predict every token.
We ask whether a model trained on every token learns to answer two-hop questions whose intermediate entity appears in no two-hop training example.
We build a synthetic corpus with such questions.
On this corpus every term of the loss can be computed exactly.
A looped transformer, a standard transformer and DiscoLoop answer almost none of these questions.
We therefore introduce the Baton family of looped transformers.
They pass the intermediate entity to the second loop as a token embedding.
When trained to predict only the answer, two of its three variants answer these questions almost perfectly.
We then relate the success and failure of each model to three design conditions.
Experiments that change one condition at a time agree with this hypothesis.
When Baton is trained on every token, however, its accuracy on these questions rises and then declines.
We further split the gradient of the next-token loss into three terms.
Removing the two terms that depend on the particular training set raises the peak accuracy but does not remove the decline.
The decline arises in the second hop while the model still recalls both facts.

%% file: sections/intro.tex
\section{Introduction}\label{sec:intro}

A language model stores facts such as ``the mother of $a$ is $b$'' and ``the birthplace of $b$ is $c$''.
Composing the two facts answers ``the birthplace of the mother of $a$'' in one forward pass without writing out the intermediate entity $b$.
We call $b$ the bridge and the task two-hop composition.
\citet{wang2024grokked} train transformers on atomic facts and on two-hop sentences that compose them.
Only some entities serve as bridges in these two-hop sentences.
After long training the models answer two-hop questions whose bridge served as a bridge in training.
They fail on the questions whose bridge never did.
We call these questions OOD questions.
A model generalizes OOD when it answers OOD questions (\cref{sec:setting}).
\citet{ye2025implicit} find a similar limit for the second hop.
Pretrained language models also struggle with latent two-hop questions \citep{yang2024latently,biran2024hopping,balesni2024lessons,karmim2026multihop}.

Controlled studies of two-hop composition train with the answer-only loss \citep{wang2024grokked,ye2025implicit,fu2026discoloop}.
The model is trained to predict only the answer of each sentence.
Language models are trained with the next-token loss on every token.
This loss gives every position a target of its own.
Many of these targets are not determined by their context.
After ``the mother of $a$'s'' any relation of the mother of $a$ can follow.
The frequencies of these relations in the training set depend on which sentences it happens to contain.
The answer-only loss thus differs from the loss of language models at every position except the answer.
Findings obtained under the answer-only loss need not hold under language-model training.
We ask whether OOD generalization appears under next-token training and whether it is kept.

We study this question in a controlled setting.
Our training set is generated from a known world of entities and relations.
Its sentences follow a fixed format.
Each term of the next-token loss can therefore be computed exactly.
We compare with a looped transformer, an eight-layer transformer and \bdisco{} \citep{fu2026discoloop}.
None of them answers more than 1.2\% of the OOD questions under either loss.
We therefore build models that generalize OOD in this setting.
We use them to study the next-token loss.
Our contributions are three.

\begin{itemize}
\item \textbf{The \archfam{} architecture family} (\cref{sec:architecture}).
These looped transformers hand the bridge to loop 2 as a token embedding.
They generalize OOD under the answer-only loss.
Under the next-token loss their OOD accuracy rises and then declines.

\item \textbf{A design hypothesis} (\cref{sec:hypothesis}).
Three conditions relate the design of a model to OOD generalization.
Every failure we observed with absolute positions violates one of them.
Experiments that change one condition at a time agree with the hypothesis.

\item \textbf{Analysis of the next-token loss} (\cref{sec:nexttoken}).
We split the gradient of the next-token loss into three terms.
Changing the targets removes two of them.
This raises the mean peak OOD accuracy from 0.78 to 0.93, but the decline of OOD accuracy remains.
During the decline the model still recalls both facts.
The decline happens in the second hop.
\end{itemize}

%% file: sections/setting.tex
\section{Setting}\label{sec:setting}

\paragraph{World.}
We use the two-graph task of \citet{fu2026discoloop}.
It follows the two-hop setting of \citet{wang2024grokked}.
The task has two disjoint graphs \GA{} and \GB{} with 500 entities each and a shared set of 50 relations.
Every entity has 10 outgoing edges with different relations.
Each edge points to another entity of the same graph.
An entity and a relation therefore determine the object.
An entity and two relations determine the end of a two-hop chain.
Each graph has 5{,}000 atomic facts $(a, r, b)$ and 50{,}000 two-hop chains $(a, r_1, r_2, c)$.
The atomic facts of a chain are $(a, r_1, b)$ and $(b, r_2, c)$.
We call $b$ the bridge.

\paragraph{Mini-language.}
We write facts as sentences of a small language.
Training on every position then resembles language-model training.
Every entity and every relation is one token.
There are three function words (\tok{'s}, \tok{is} and \tok{.}).
An atomic fact is the sentence $a$~\tok{'s}~$r$~\tok{is}~$b$~\tok{.} (\cref{tab:format}).
A two-hop chain is the sentence $a$~\tok{'s}~$r_1$~\tok{'s}~$r_2$~\tok{is}~$c$~\tok{.}.
Two properties of this format are used throughout the paper.
First, a two-hop sentence describes its bridge as $a$~\tok{'s}~$r_1$ but never names it.
Second, the two sentence types share this description and differ in the token after it.
An atomic sentence continues with \tok{is} and then $b$.
A two-hop sentence continues with \tok{'s} and then a relation of $b$.
We call the position of this token the position after the description.
In both types the answer is the token after \tok{is}.

\begin{table}[t]
\caption{The two sentence types aligned by position.
Both begin with a description $a$~\tok{'s}~$r$ of an entity $b$.
The shaded token follows the description.}
\label{tab:format}
\centering
\small
\begin{tabular}{@{}lcccccccc@{}}
\toprule
atomic  & $a$ & \tok{'s} & $r$   & \shade{\tok{is}} & $b$   & \tok{.}  &     &         \\
two-hop & $a$ & \tok{'s} & $r_1$ & \shade{\tok{'s}} & $r_2$ & \tok{is} & $c$ & \tok{.} \\
\cmidrule(lr){2-4}
        & \multicolumn{3}{c}{description of $b$} & \multicolumn{5}{l}{$\uparrow$ after the description} \\
\bottomrule
\end{tabular}
\end{table}

\paragraph{Training and test sets.}
The training set contains all 10{,}000 atomic facts of both graphs.
It also contains 10{,}000 two-hop sentences of \GA{} drawn at random from its 50{,}000 chains.
A test question is a two-hop sentence cut after \tok{is} ($a$~\tok{'s}~$r_1$~\tok{'s}~$r_2$~\tok{is}).
The model must predict its answer $c$ as the next token.
ID accuracy is measured on 2{,}000 other two-hop questions of \GA{}.
OOD accuracy is measured on 2{,}000 two-hop questions of \GB{}.
The atomic facts of \GB{} are trained.
No \GB{} entity appears in a training two-hop sentence, so no \GB{} entity serves as a bridge in training.
Following \citet{varma2023grokking} we distinguish two kinds of solution that fit this training set.
The general composition \Cgen{} answers a question by recalling two facts in turn.
It recalls the bridge from its description and then the answer from the bridge and $r_2$.
It uses the same computations for every entity.
Solutions of the second kind (\Cmem{}) answer from what the two-hop training sentences contained.
Examples are stored answers or a second hop that works only for entities that served as bridges.
\Cmem{} can answer ID questions but not OOD questions.
OOD accuracy therefore measures \Cgen{} alone.

\paragraph{Losses.}
The next-token loss is the cross-entropy of every token after the first, averaged over all predicted tokens.
The final \tok{.} is included.
It is the loss of language-model training.
The answer-only loss is the cross-entropy of the prediction at \tok{is} alone.
Its target is the answer of the sentence.
The two losses differ only in which positions carry a target.
\Cref{sec:nexttoken} introduces variants of the next-token loss that change its targets at some positions.

\paragraph{Models and training.}
All models share one backbone of pre-LN transformer blocks with width 256 and 4 attention heads.
Learned absolute position vectors are added to the input.
We use them because answer-only training then keeps its OOD accuracy in every seed.
The comparison of the two losses needs such a stable reference (\cref{app:rotary}).
The output layer is tied to the input embedding.
\Cref{sec:nexttoken} and \cref{app:rotary} also use rotary position embeddings (base $10^4$) instead of absolute vectors.
A looped model applies the same 4 blocks twice and reads the logits after the second pass.
We compare with three baselines.
\bloop{} is this looped model without further changes.
\btf{} stacks 8 distinct blocks without a loop.
It has the same compute per token and twice the parameters (6.6M against 3.4M).
\bdisco{} \citep{fu2026discoloop} adds re-entry at every position.
Between the two passes it maps the predicted next-token distribution of each position back through the embedding and adds it to the state of that position.
The Baton variants (\cref{sec:architecture}) add fewer than 3{,}000 parameters to \bloop{}.
All models are trained with AdamW for 3{,}000 epochs with a cosine decay of the learning rate.
Each configuration is run with four or six seeds.
\Cref{app:config} gives all settings.

\paragraph{Evaluation.}
A question counts as answered when the argmax prediction at \tok{is} is its answer.
We measure ID and OOD accuracy every 20 epochs.
We say that a run generalizes OOD when its OOD accuracy exceeds 0.1 at some point in training.
The highest OOD accuracy of every run we report is either at most 0.08 or at least 0.45.
The choice of this threshold therefore does not matter.
OOD accuracy can rise early in training and fall late.
We therefore report two numbers per run and plot moving means.
The moving mean of a run is its OOD accuracy averaged over a window of 250 epochs.
The peak is the highest value of the moving mean.
The final level is the mean OOD accuracy over the last 500 epochs (the last 25 measurements).
\Cref{app:seeds} gives the results of every seed.

%% file: sections/architecture.tex
\section{The Baton architecture}\label{sec:architecture}

\begin{figure*}[t]
\centering
\begin{minipage}[c]{0.66\textwidth}
\centering
\input{figures/arch}
\end{minipage}\hfill
\begin{minipage}[c]{0.31\textwidth}
\caption{The Baton family.
(a) The streams and the two loops shared by all variants.
(b)--(d) The state that writes the bridge $b$ into its slot in each variant and the answer position that reads it in loop 2 (dashed).
Frame colours as in \cref{fig:ood}.}
\label{fig:arch}
\end{minipage}
\end{figure*}
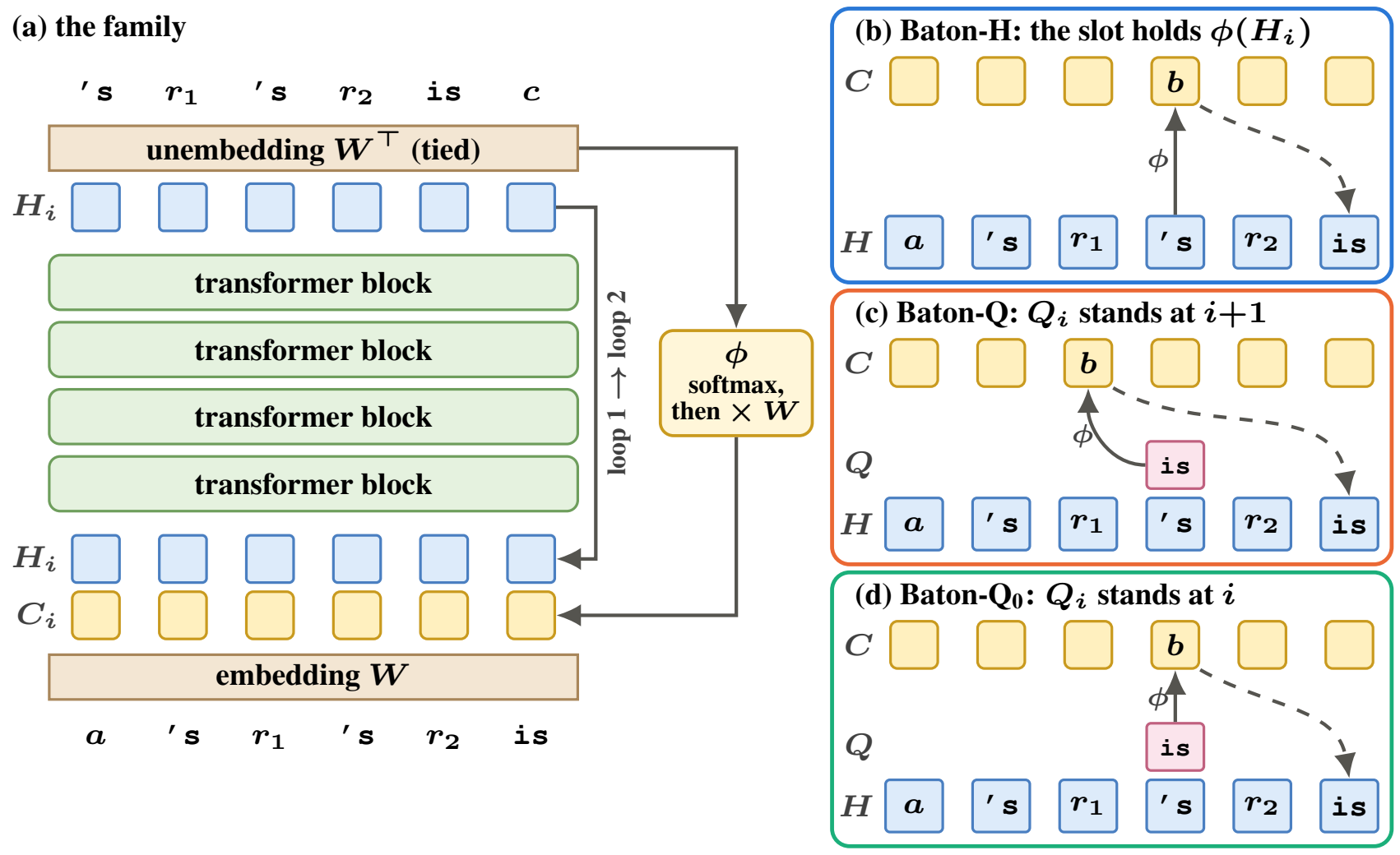

Baton is a family of looped models that hand the result of loop 1 to loop 2 as a token embedding.
We use it as an instrument for the question of this paper.
Under the answer-only loss its variants generalize to OOD questions in our setting.
The three baselines do not.
This section describes the three variants and their OOD accuracy under the answer-only loss and the next-token loss.
\Cref{sec:hypothesis} relates their results to the design.

\subsection{Design}\label{sec:design}

\paragraph{Structure.}
\Cref{fig:arch} shows the family.
A looped model runs the same blocks twice.
We call the two passes loop 1 and loop 2.
Each position $i$ has a state $H_i$.
The output layer reads the logits at position $i$ from $H_i$.
Every variant adds a stream of slots $C$ with one slot $C_i$ per position.
\mq{} and \mqal{} add a third stream $Q$ with one state $Q_i$ per position.
All streams pass through the same blocks.
The states $H_i$ and $Q_i$ carry over from loop 1 to loop 2.
Only the logits read from the states $H_i$ enter the loss.
$H_i$ receives the position vector $p_i$.
$C_i$ receives its own learned vector $s_i$.

\paragraph{Attention.}
We write $\mathcal{K}(X)$ for the set of states that a state $X$ attends to in every block.
In every variant
\begin{equation}
  \mathcal{K}(H_i) = \{H_j : j \le i\} \cup \{C_j : j < i\}.
  \label{eq:readH}
\end{equation}
In \mq{} and \mqal{} each slot attends only to itself ($\mathcal{K}(C_i) = \{C_i\}$) and passes the feed-forward layers.
In \mbase{} the slots neither attend nor pass the feed-forward layers.
They only supply keys and values.
The stream $Q$ of \mq{} and \mqal{} attends as
\begin{equation}
  \mathcal{K}(Q_i) =
  \begin{cases}
    \{Q_i\} \cup \{H_j, C_j : j \le i\} & \text{in \mq{}}, \\
    \{Q_i\} \cup \{H_j, C_j : j < i\}   & \text{in \mqal{}}.
  \end{cases}
  \label{eq:readQ}
\end{equation}
$Q_i$ receives the position vector $p_{i+1}$ in \mq{} and $p_i$ in \mqal{}.

\paragraph{Slots $C$.}
$C_i$ holds the embedding of the input token $x_i$ at position $i$ in loop 1.
Between the two loops $C_i$ is replaced by the decode of a state $h$
\begin{equation}
  \phi(h) = \E_{v \sim \qmodel(\cdot \mid h)} \big[ W_v \big].
  \label{eq:phi}
\end{equation}
Here $W$ is the tied embedding matrix and $W_v$ is the embedding of token $v$.
$\qmodel(\cdot \mid h)$ is the next-token distribution that the output layer reads from $h$.
$C_i$ holds approximately $W_b$ when this distribution predicts the bridge $b$.
The decode is the re-entry of \citet{fu2026discoloop}.
\bdisco{} adds it back to the state it was decoded from.
Baton writes it into $C_i$.

\paragraph{Stream $Q$.}
The variants differ in which state is decoded into $C_i$.
\mbase{} writes $\phi(H_i)$ into $C_i$.
\mq{} and \mqal{} write $\phi(Q_i)$ into $C_i$.
The input of every $Q_i$ is the embedding of \tok{is} plus a vector $\delta$ learned from zero.
$Q$ has no loss.
$H$ does not read $Q$.
\mq{} and \mqal{} differ only in the position vector of $Q_i$ and in the positions that $Q_i$ reads (\cref{eq:readQ}).

\subsection{Results}\label{sec:arch-results}
\begin{figure*}[t]
\centering
\includegraphics[width=\textwidth]{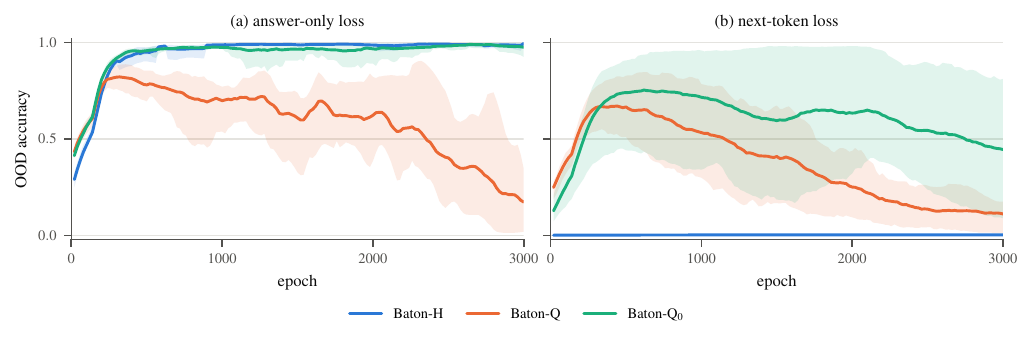}
\caption{Moving mean of the OOD accuracy of the three variants.
Lines show the mean over seeds and bands show the range.}
\label{fig:ood}
\end{figure*}
\begin{table}[t]
\caption{Peak and final level of the OOD accuracy.
Mean over seeds with the range in parentheses.}
\label{tab:ood}
\centering
\footnotesize
\setlength{\tabcolsep}{3pt}
\input{tables/ood_variants}
\end{table}

\paragraph{Baselines.}
No run of the three baselines generalizes OOD under the answer-only loss or the next-token loss.
Their OOD accuracy stays at or below 0.012 at every evaluation.
Their ID accuracy stays below 0.32.
\bloop{} under the answer-only loss reaches the highest ID accuracy (final level 0.16--0.25).
\citet{fu2026discoloop} report that \bdisco{} generalizes OOD in their setting (\cref{app:disco}).
\Cref{fig:ood} does not show the baselines.

\paragraph{Answer-only loss.}
Every run of the three variants generalizes OOD under the answer-only loss (\cref{fig:ood}a and \cref{tab:ood}).
\mbase{} and \mqal{} reach 0.98--1.00 and keep it (final level 0.96--1.00 in every seed).
\mq{} peaks at 0.78--0.91 and then declines to a final level of 0.06--0.56.

\paragraph{Next-token loss.}
No run of \mbase{} generalizes OOD under the next-token loss (\cref{fig:ood}b and \cref{tab:ood}).
Its OOD accuracy stays at or below 0.01.
Every run of \mqal{} generalizes OOD.
In every seed its final level is below that of answer-only training (0.14--0.86 against 0.96--1.00).
In four seeds it peaks at 0.89--0.98 and then declines.
In the other two it peaks at about 0.5.
\mq{} peaks at 0.55--0.85 and ends at 0.02--0.23.

\Cref{sec:hypothesis} relates these differences between the variants to their design.
\Cref{sec:nexttoken} studies how the next-token loss lowers the final level of \mqal{}.

%% file: figures/arch.tex
\definecolor{archH}{HTML}{DDEBFA}\definecolor{archHd}{HTML}{4A7FC1}
\definecolor{archQ}{HTML}{FBE4EC}\definecolor{archQd}{HTML}{C0607F}
\definecolor{archC}{HTML}{FFF0BF}\definecolor{archCd}{HTML}{C99A1E}
\definecolor{archEmb}{HTML}{F3E6D3}\definecolor{archEmbd}{HTML}{A8865A}
\definecolor{archBlk}{HTML}{E5F2DF}\definecolor{archBlkd}{HTML}{6E9E5C}\definecolor{archLine}{HTML}{55544F}
\def\dx{0.7}
\pgfmathsetmacro{\W}{5*\dx}%
\pgfmathsetmacro{\mid}{2.5*\dx}%
\pgfmathsetmacro{\barw}{5*\dx+0.75}%
\begin{tikzpicture}[
  x=1cm, y=1.08cm, font=\scriptsize\bfseries\boldmath,
  cell/.style={minimum width=3.8mm, minimum height=3.8mm, inner sep=0pt, line width=0.6pt, rounded corners=1pt},
  hcell/.style={cell, draw=archHd, fill=archH},
  qcell/.style={cell, draw=archQd, fill=archQ},
  ccell/.style={cell, draw=archCd, fill=archC, rounded corners=1.6pt},
  bar/.style={draw=archEmbd, fill=archEmb, line width=0.6pt, minimum height=3.6mm, inner sep=0pt, minimum width=\barw cm},
  blk/.style={bar, draw=archBlkd, fill=archBlk, rounded corners=2.5pt, minimum height=4.4mm},
  arr/.style={-{Latex[length=2mm, width=1.8mm]}, archLine, line width=0.8pt},
  tk/.style={inner sep=0pt, text height=1.4ex, text depth=0.3ex},
  lab/.style={inner sep=1pt, text=black!75},
]
\begin{scope}
  \node[anchor=west] at (-0.8, 5.25) {\textbf{(a)} the family};
  \foreach \t [count=\k from 0] in {$a$, \tok{'s}, $r_1$, \tok{'s}, $r_2$, \tok{is}}
    \node[tk] at (\k*\dx, 0) {\t};
  \node[bar] at (\mid, 0.42) {embedding $W$};
  \foreach \k in {0,...,5} \node[ccell] (ci\k) at (\k*\dx, 0.88) {};
  \foreach \k in {0,...,5} \node[hcell] (hi\k) at (\k*\dx, 1.30) {};
  \foreach \y in {1.86, 2.36, 2.86, 3.36} \node[blk] at (\mid, \y) {transformer block};
  \foreach \k in {0,...,5} \node[hcell] (ho\k) at (\k*\dx, 3.92) {};
  \node[bar] (unemb) at (\mid, 4.36) {unembedding $W^{\top}$ (tied)};
  \foreach \t [count=\k from 0] in {\tok{'s}, $r_1$, \tok{'s}, $r_2$, \tok{is}, $c$}
    \node[tk] at (\k*\dx, 4.80) {\t};
  \node[lab, anchor=east] at (-0.25, 0.88) {$C_i$};
  \node[lab, anchor=east] at (-0.25, 1.30) {$H_i$};
  \node[lab, anchor=east] at (-0.25, 3.92) {$H_i$};
  \draw[arr] (ho5.east) -- (\W+0.52, 3.92) -- (\W+0.52, 1.30) -- (hi5.east);
  \node[lab, rotate=90, anchor=north] at (\W+0.57, 2.61) {\tiny loop 1 $\to$ loop 2};
  \node[draw=archCd, line width=0.7pt, rounded corners=3pt, fill=archC!60, align=center, inner sep=2.5pt] (phi)
    at (\W+1.65, 2.61) {$\phi$\\[-1pt]{\tiny softmax,}\\[-2pt]{\tiny then ${\times}\,W$}};
  \draw[arr] (unemb.east) -- (\W+1.65, 4.36) -- (phi.north);
  \draw[arr] (phi.south) -- (\W+1.65, 0.88) -- (ci5.east);
\end{scope}
\pgfmathsetmacro{\X}{\W+3.08}
\newcommand{\archpanel}[7]{%
  \begin{scope}[shift={(\X, #1)}]
    \draw[#2, line width=0.9pt, rounded corners=4pt] (-0.66, -0.08) rectangle (\W+0.34, 1.96);
    \node[anchor=west] at (-0.6, 1.78) {#3};
    \foreach \t [count=\k from 0] in {$a$, \tok{'s}, $r_1$, \tok{'s}, $r_2$, \tok{is}}
      \node[hcell, minimum width=4.6mm, minimum height=4.2mm] (h\k) at (\k*\dx, 0.22) {\t};
    \foreach \k in {0,...,5} \node[ccell] (c\k) at (\k*\dx, 1.42) {};
    \node[lab, anchor=east] at (-0.28, 0.22) {$H$};
    \node[lab, anchor=east] at (-0.28, 1.42) {$C$};
    \node at (c#4) {$b$};
    \ifx h#6
      \draw[arr] (h#5.north) -- node[lab, left, pos=0.5] {\tiny$\phi$} (c#4.south);
    \else
      \node[lab, anchor=east] at (-0.28, 0.66) {$Q$};
      \node[qcell, minimum width=4.6mm] (q) at (#5*\dx, 0.66) {\tiny\texttt{is}};
      \ifx s#7
        \draw[arr] (q.north) -- node[lab, left, pos=0.45] {\tiny$\phi$} (c#4.south);
      \else
        \draw[arr] (q.west) to[out=180, in=-90] node[lab, left, pos=0.55] {\tiny$\phi$} (c#4.south);
      \fi
    \fi
    \draw[arr, dashed] (c#4.south east) to[out=-30, in=100] (h5.north);
  \end{scope}}
\archpanel{3.44}{vMtwo}{\textbf{(b)} \mbase{}: the slot holds $\phi(H_i)$}{3}{3}{h}{s}
\archpanel{1.34}{vMeight}{\textbf{(c)} \mq{}: $Q_i$ stands at $i{+}1$}{2}{3}{q}{c}
\archpanel{-0.76}{vMeightal}{\textbf{(d)} \mqal{}: $Q_i$ stands at $i$}{3}{3}{q}{s}
\end{tikzpicture}

%% file: tables/ood_variants.tex
\begin{tabular}{@{}llccc@{}}
\toprule
loss & variant & seeds & peak & final level \\
\midrule
answer-only & \mbase{} & 4 & 1.00\,{\scriptsize(0.99--1.00)} & 0.99\,{\scriptsize(0.98--1.00)} \\
 & \mq{} & 4 & 0.85\,{\scriptsize(0.78--0.91)} & 0.29\,{\scriptsize(0.06--0.56)} \\
 & \mqal{} & 6 & 0.99\,{\scriptsize(0.98--1.00)} & 0.99\,{\scriptsize(0.96--1.00)} \\
\midrule
next-token & \mbase{} & 4 & 0.00\,{\scriptsize(0.00--0.00)} & 0.00\,{\scriptsize(0.00--0.00)} \\
 & \mq{} & 4 & 0.69\,{\scriptsize(0.55--0.85)} & 0.12\,{\scriptsize(0.02--0.23)} \\
 & \mqal{} & 6 & 0.78\,{\scriptsize(0.50--0.98)} & 0.49\,{\scriptsize(0.14--0.86)} \\
\bottomrule
\end{tabular}

%% file: sections/hypothesis.tex
\section{A design hypothesis}\label{sec:hypothesis}

The models of \cref{sec:architecture} differ in whether they generalize OOD and in whether they keep the OOD accuracy.
We relate these differences to three conditions.
Every failure we observed with absolute positions violates one of them.
Three tests change one condition each.
Their outcomes agree with the hypothesis.

\subsection{Statement}\label{sec:statement}

\begin{quote}
OOD generalization in two-hop composition is supported when three conditions hold.\par
\textbf{Condition 1.} Both hops run the computation that answers atomic facts.\par
\textbf{Condition 2.} The bridge enters the second hop as its token embedding, the form in which an entity enters an atomic sentence.\par
\textbf{Condition 3.} The answer position cannot read a decode of the answer that the model computed in loop 1.
\end{quote}

A decode of the loop-1 answer lets the answer position copy that answer.
A training sentence answered by such a copy no longer trains the computation of Condition~1.
No two-hop training sentence uses a \GB{} entity as its bridge.
Only atomic sentences therefore train this computation for \GB{} entities.

We say that a model fails when it never generalizes OOD or when its OOD accuracy declines under the answer-only loss.
A decline under the next-token loss is the subject of \cref{sec:nexttoken}.
With rotary positions answer-only training of \mbase{} and \mqal{} also declines (\cref{app:rotary}).
The hypothesis does not explain this decline.
The hypothesis comes from the design of Baton and from the failures we observed.
\Cref{sec:conditions} relates each condition to both.

\subsection{The conditions in the observed models}\label{sec:conditions}

\begin{table}[t]
\caption{A condition that each model violates and the course of its OOD accuracy.
Rows that name no loss hold for both losses.}
\label{tab:conditions}
\centering
\small
\begin{tabular}{@{}lcl@{}}
\toprule
model & condition & OOD accuracy \\
\midrule
\btf{}                & 1    & never above 0.012 \\
\bloop{}              & 2    & never above 0.012 \\
\mbase{}, next-token  & 1    & never above 0.012 \\
\bdisco{}             & 3    & never above 0.012 \\
\mq{}                 & 3    & rises and declines \\
\mbase{}, answer-only & none & rises and stays \\
\mqal{}, answer-only  & none & rises and stays \\
\mqal{}, next-token   & none & rises and ends lower \\
\bottomrule
\end{tabular}
\end{table}

\begin{figure}[t]
\centering
\input{figures/duplicate}
\caption{The duplicate slot in an atomic sentence $a$~\tok{'s}~$r$~\tok{is}~$b$ (positions 1 to 5).
The $Q$ with the position vector $p_4$ of the answer position computes the answer $b$ in loop 1.
(a) In \mq{} this $Q$ is $Q_3$ and writes $C_3$.
$H_4$ reads $C_3$ in loop 2 (dashed).
(b) In \mqal{} this $Q$ is $Q_4$ and writes $C_4$.
$H_4$ does not read $C_4$.}
\label{fig:duplicate}
\end{figure}
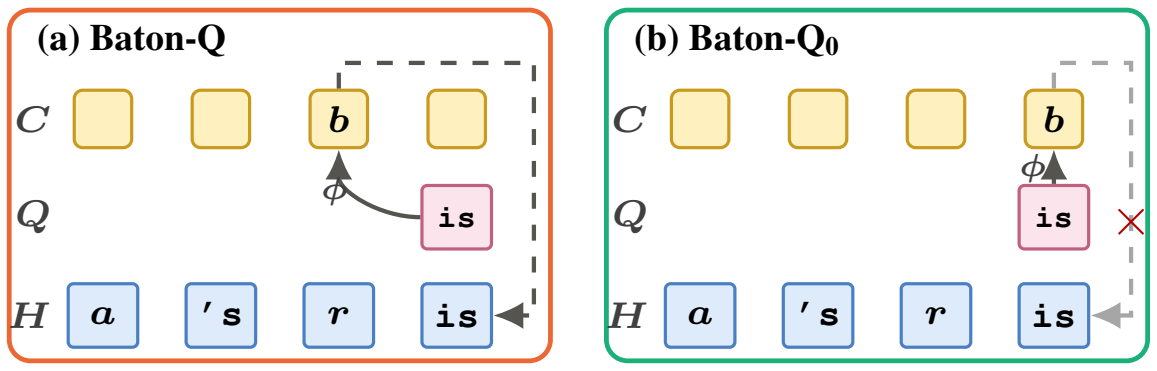

\Cref{tab:conditions} lists a condition that each model violates.
We number the positions of a sentence from 1.
The answer position \tok{is} is then position~6 in a two-hop sentence and position~4 in an atomic sentence.

\paragraph{Condition 1.}
Both loops of Baton apply the same blocks.
In an atomic sentence the state $H_4$ at \tok{is} reads the subject $a$ and the relation $r$.
It predicts the answer $b$.
In a two-hop sentence the state $H_6$ at \tok{is} reads the bridge from the slot $C_4$ in loop 2 ($C_3$ in \mq{}).
It also reads the relation $r_2$ and predicts the answer $c$.
$H_4$ and $H_6$ have the same input token \tok{is} and run the same blocks.
The second hop therefore runs the computation that answers atomic facts.

The first hop computes the bridge $b$ from the description $a$~\tok{'s}~$r_1$ in loop 1.
In an atomic sentence $H_4$ computes the answer from a description of the same form.
In \mbase{} the bridge is computed by $H_4$ of the two-hop sentence.
Under the next-token loss the target of this $H_4$ is the token $r_2$.
$H_4$ is then trained to predict a relation of the bridge and not the bridge.
\mbase{} therefore violates Condition~1 under the next-token loss.
Its first hop does not run the computation that answers atomic facts.
Under the answer-only loss this $H_4$ has no target.

As for \mq{} and \mqal{}, the bridge is computed in the stream $Q$.
The $Q$ that computes it ($Q_3$ in \mq{} and $Q_4$ in \mqal{}) has the position vector $p_4$.
It reads the description $a$~\tok{'s}~$r_1$ (\cref{eq:readQ}).
Its input is the embedding of \tok{is} plus a vector $\delta$ learned from zero (\cref{sec:design}).
At initialization this $Q$ therefore has the input and the position vector of $H_4$ in an atomic sentence.
This choice biases the first hop toward the computation that answers atomic facts and thus toward Condition~1.
$Q$ has no loss.
No target pulls the first hop away from this computation.

\btf{} violates Condition~1.
Its 8 blocks are distinct.
The second hop cannot reuse the computation of the first hop and needs its own copy of each fact.
\citet{wang2024grokked} trace the OOD failure of non-looped transformers to this.

\paragraph{Condition 2.}
We call the state whose decode fills a slot the writer of that slot.
In \mbase{} $H_i$ is the writer of $C_i$.
In \mq{} and \mqal{} $Q_i$ is the writer of $C_i$ (\cref{fig:arch}b--d).
The decode $\phi$ (\cref{eq:phi}) writes the next-token prediction of the writer into the slot as an embedding.
When the writer predicts the bridge $b$, the slot holds approximately $W_b$.
This is the embedding that the model learns for $b$ in atomic sentences.
Every Baton variant thus satisfies Condition~2 through the decode.
Whether the slot holds the bridge depends on the first hop (Condition~1).

\bloop{} violates Condition~2.
Loop 2 receives the bridge only inside a hidden state.
With many more two-hop facts per atomic fact a looped model can generalize OOD (\cref{sec:related}).
This account of the failure of \bloop{} holds for the two or four two-hop sentences per atomic fact that we train on.

\paragraph{Condition 3.}
$H_i$ reads only the slots $C_j$ of earlier positions $j < i$ (\cref{eq:readH}).
In \mbase{} and \mqal{} the answer position therefore cannot read a decode of its loop-1 answer.
\bdisco{} violates Condition~3.
Between the loops it adds the decode of each state back to that state.
At the answer position loop 2 then receives the decode of the loop-1 answer and can copy it.
Without this re-entry at the answer position \bdisco{} generalizes OOD in the symbolic format of \citet{fu2026discoloop} (OOD accuracy 0.97 and 1.00 in two seeds, \cref{app:disco}).

\mq{} violates Condition~3 through a duplicate slot (\cref{fig:duplicate}).
Its $Q_i$ has the position vector $p_{i+1}$ and writes $C_i$.
\mq{} thus places the bridge in $C_3$ at the position of $r_1$.
The tokens \tok{'s}, $r_2$ and \tok{is} then follow the bridge at the distances at which they follow the subject of an atomic sentence.
The same placement acts in atomic sentences.
In an atomic sentence $Q_3$ has the position vector $p_4$ and computes the answer $b$ in loop 1.
It writes $b$ into $C_3$.
$H_4$ reads $C_3$ in loop 2 and can copy $b$.
We call $C_3$ the duplicate slot.
The atomic sentences of \GB{} then need not keep training the computation of Condition~1 for \GB{} entities.
Yet the second hop of an OOD question runs exactly this computation.
\Cref{app:duplicate} measures how many atomic answers follow the duplicate slot.
\mqal{} gives $Q_i$ the position vector $p_i$.
In an atomic sentence $Q_4$ then computes $b$ and writes $C_4$.
$H_4$ does not read $C_4$.
\mqal{} thereby gives up the distances of an atomic sentence.

\subsection{Tests that change one condition}\label{sec:tests}

\begin{table*}[t]
\caption{OOD accuracy when one condition changes (four seeds, mean with the range in parentheses).
Each test follows the variant that it changes.
The column \emph{condition} names the condition that the change restores or breaks.}
\label{tab:tests}
\centering
\small
\input{tables/tests}
\end{table*}

Each test changes one condition of one Baton variant and keeps everything else (\cref{tab:tests}).
Two tests restore a condition that a variant violates.
The third test breaks a condition that a variant satisfies.

\paragraph{Writer's target removed.}
In two-hop sentences we remove the target token of $H_4$ from the next-token loss of \mbase{}.
This target is the relation $r_2$.
All other targets stay.
All four seeds then generalize OOD (peak 0.81--0.98 against 0.00).
ID accuracy rises from 0.02--0.04 to 0.99--1.00.
OOD accuracy declines afterwards to a final level of 0.01--0.13.

\paragraph{Duplicate slot hidden.}
In loop 2 we hide $C_{i-1}$ from the state $H_i$ at \tok{is} in \mq{}.
In an atomic sentence this hides the duplicate slot $C_3$ from $H_4$.
The final level rises in all four seeds (0.44--0.95 against 0.06--0.56).
Three seeds end at 0.88--0.95.
In the fourth seed OOD accuracy still declines from 0.97 to 0.44.

\paragraph{Undecoded slot.}
We predicted the outcome of this test from the hypothesis before running it.
In \mqal{} we write into $C_i$ the state $Q_i$ after the final LayerNorm instead of $\phi(Q_i)$.
We rescale this state to the mean norm of the token embeddings.
The slot then holds the bridge as a hidden state and not as its token embedding.
Condition~2 predicts no OOD generalization.
No seed generalizes OOD under either loss (final level at most 0.03).
Under the answer-only loss ID accuracy reaches 0.92--0.94.
$Q_4$ still predicts the correct bridge for at least 0.999 of the OOD questions.
The failure therefore lies in the form of the input that loop 2 receives.

\paragraph{The next-token loss.}
Two models in this section satisfy all three conditions by design under the next-token loss.
These are \mqal{} and \mbase{} without the target token of $H_4$.
Both generalize OOD in every seed.
Both end below the answer-only training of the same variant in every seed (\cref{tab:ood,tab:tests}).
The hypothesis does not explain this decline.
During the decline of \mqal{} the first hop still writes the correct bridge into its slot.
The atomic facts of \GB{} are still answered (\cref{sec:where}).
\Cref{sec:nexttoken} analyses this decline.

%% file: figures/duplicate.tex
\definecolor{archH}{HTML}{DDEBFA}\definecolor{archHd}{HTML}{4A7FC1}
\definecolor{archQ}{HTML}{FBE4EC}\definecolor{archQd}{HTML}{C0607F}
\definecolor{archC}{HTML}{FFF0BF}\definecolor{archCd}{HTML}{C99A1E}
\definecolor{archLine}{HTML}{55544F}
\def\ddx{0.78}
\begin{tikzpicture}[x=1cm, y=1cm, font=\scriptsize\bfseries\boldmath,
  cell/.style={minimum width=4.6mm, minimum height=4.2mm, inner sep=0pt, line width=0.6pt, rounded corners=1pt},
  hcell/.style={cell, draw=archHd, fill=archH},
  qcell/.style={cell, draw=archQd, fill=archQ},
  ccell/.style={cell, draw=archCd, fill=archC, rounded corners=1.6pt, minimum width=3.8mm, minimum height=3.8mm},
  arr/.style={-{Latex[length=2mm, width=1.8mm]}, archLine, line width=0.8pt},
  lab/.style={inner sep=1pt, text=black!75},
]
\newcommand{\duppanel}[5]{%
  \begin{scope}[shift={(#1, 0)}]
    \draw[#2, line width=0.9pt, rounded corners=4pt] (-0.62, -0.30) rectangle (3*\ddx+0.62, 2.02);
    \node[anchor=west] at (-0.56, 1.82) {#3};
    \foreach \t [count=\k from 0] in {$a$, \tok{'s}, $r$, \tok{is}}
      \node[hcell] (h\k) at (\k*\ddx, 0.0) {\t};
    \foreach \k in {0,...,3} \node[ccell] (c\k) at (\k*\ddx, 1.30) {};
    \node[lab, anchor=east] at (-0.30, 0.0) {$H$};
    \node[lab, anchor=east] at (-0.30, 0.65) {$Q$};
    \node[lab, anchor=east] at (-0.30, 1.30) {$C$};
    \node[qcell] (q) at (3*\ddx, 0.65) {\tiny\texttt{is}};
    \node at (c#4) {$b$};
    \ifnum#4=3
      \draw[arr] (q.north) -- node[lab, left, pos=0.45] {\tiny$\phi$} (c#4.south);
    \else
      \draw[arr] (q.west) to[out=180, in=-90] node[lab, left, pos=0.6] {\tiny$\phi$} (c#4.south);
    \fi
    \ifx y#5
      \draw[arr, dashed] (c#4.north) -- ++(0, 0.17) -| ($(h3.east)+(0.27, 0)$) -- (h3.east);
    \else
      \draw[arr, dashed, black!35] (c#4.north) -- ++(0, 0.17) -| ($(h3.east)+(0.27, 0)$) -- (h3.east);
      \node[text=red!70!black, font=\normalsize] at ($(h3.east)+(0.27, 0.62)$) {$\times$};
    \fi
  \end{scope}}
\duppanel{0}{vMeight}{\textbf{(a)} \mq{}}{2}{y}
\duppanel{3.95}{vMeightal}{\textbf{(b)} \mqal{}}{3}{n}
\end{tikzpicture}

%% file: tables/tests.tex
\begin{tabular}{@{}lllccc@{}}
\toprule
model & change & condition & peak & final level & ID final level \\
\midrule
\mbase{}, next-token & -- & -- & 0.00\,{\scriptsize(0.00--0.00)} & 0.00\,{\scriptsize(0.00--0.00)} & 0.03\,{\scriptsize(0.02--0.04)} \\
 & writer's target removed & restores 1 & 0.93\,{\scriptsize(0.81--0.98)} & 0.06\,{\scriptsize(0.01--0.13)} & 1.00\,{\scriptsize(0.99--1.00)} \\
\midrule
\mq{}, answer-only & -- & -- & 0.85\,{\scriptsize(0.78--0.91)} & 0.29\,{\scriptsize(0.06--0.56)} & 1.00\,{\scriptsize(1.00--1.00)} \\
 & duplicate slot hidden in loop 2 & restores 3 & 0.98\,{\scriptsize(0.96--0.99)} & 0.80\,{\scriptsize(0.44--0.95)} & 1.00\,{\scriptsize(1.00--1.00)} \\
\midrule
\mqal{}, answer-only & -- & -- & 1.00\,{\scriptsize(0.99--1.00)} & 0.99\,{\scriptsize(0.98--1.00)} & 1.00\,{\scriptsize(1.00--1.00)} \\
 & undecoded slot & breaks 2 & 0.02\,{\scriptsize(0.02--0.03)} & 0.02\,{\scriptsize(0.01--0.03)} & 0.92\,{\scriptsize(0.92--0.94)} \\
\bottomrule
\end{tabular}

%% file: sections/decomposition.tex
\section{Analysis of the next-token loss}\label{sec:nexttoken}

Under the next-token loss the final OOD accuracy of \mqal{} is below that of answer-only training in every seed (\cref{sec:architecture}).
This section analyses which part of the next-token loss lowers it.
\Cref{sec:decomposition} splits the loss and its gradient into three terms.
\Cref{sec:ladder} removes these terms one at a time by changing the targets.
\Cref{sec:where} shows where in the model the OOD answers are lost.

\subsection{Decomposition}\label{sec:decomposition}

\paragraph{The distributions \ptrain{} and \pref{}.}
A context $x$ is the sequence of tokens of a training sentence up to some position.
We write $\ptrain(\cdot \mid x)$ for the frequencies of the next token after $x$ in the training set.
The second distribution $\pref(\cdot \mid x)$ treats all entities alike.
Take the context $a$~\tok{'s}~$r_1$~\tok{'s}.
Its next token is a relation of the bridge $b$.
In the training set only some of the ten relations of $b$ follow this prefix.
\ptrain{} gives their frequencies.
\pref{} is uniform over all ten relations of $b$.
In general \pref{} is the best fit to the training set by a predictor that treats all entities alike and uses only the facts of the two hops.
These facts are the relations of $a$, the bridge $b$ and the relations of $b$.
Such a predictor cannot tell whether $a$ belongs to \GA{} or to \GB{}.
The context determines the next token everywhere except at three positions.
After $a$~\tok{'s} the distribution \pref{} is uniform over the relations of $a$.
After $a$~\tok{'s}~$r_1$ it gives \tok{'s} and \tok{is} the probability $1/2$ each.
This is the share of descriptions followed by \tok{'s} in the training set.
After $a$~\tok{'s}~$r_1$~\tok{'s} it is uniform over the relations of $b$.
At every other position \pref{} and \ptrain{} agree.

\paragraph{Decomposition of the loss.}
Let $\qmodel(\cdot \mid x)$ be the prediction of the model.
The expected next-token loss over the training contexts splits into three terms
\begin{equation}
  \begin{aligned}
  L(\theta)
  &= \underbrace{\E_x H\big(\ptrain(\cdot \mid x)\big)}_{\Htrain}
   + \underbrace{\E_x \KL\big(\ptrain \,\|\, \pref\big)}_{\Res} \\
  &\quad + \underbrace{\E_x \E_{y \sim \ptrain} \log \frac{\pref(y \mid x)}{\qmodel(y \mid x)}}_{\Delta(\theta)} .
  \end{aligned}
  \label{eq:loss}
\end{equation}
\Htrain{} is the conditional entropy of the training set and does not depend on the model.
\Res{} does not depend on the model either.
Only $\Delta(\theta)$ depends on the model.
A model that predicts \pref{} has $\Delta = 0$.
A model that reproduces \ptrain{} has $\Delta = -\Res$ and the lowest possible loss.
\Res{} is therefore the amount by which reproducing the training frequencies lowers the loss below that of \pref{}.
Our training set comes from a known generator.
\Res{} can therefore be computed exactly (\cref{app:residual}).

\begin{table}[t]
\caption{\Res{} by context in nats per two-hop training sentence.}
\label{tab:residual-pos}
\centering
\small
\begin{tabular}{@{}lc@{}}
\toprule
context and next token & \Res{} \\
\midrule
$a$~\tok{'s} $\to$ $r_1$ & 0.12 \\
$a$~\tok{'s}~$r_1$ $\to$ \tok{'s} or \tok{is} (\GA{} entities) & 0.19 \\
$a$~\tok{'s}~$r_1$ $\to$ \tok{'s} or \tok{is} (\GB{} entities) & 0.35 \\
$a$~\tok{'s}~$r_1$~\tok{'s} $\to$ $r_2$ & 1.37 \\
\midrule
total & 2.03 \\
\bottomrule
\end{tabular}
\end{table}

\Cref{tab:residual-pos} gives \Res{} by context.
\Res{} is largest after $a$~\tok{'s}~$r_1$~\tok{'s}.
There the training set contains only some of the relations of each bridge.
\Res{} also falls on the descriptions of \GB{} entities.
The training set never continues them with \tok{'s}.
\pref{} gives \tok{'s} the probability $1/2$ there.
The split into ID and OOD questions therefore leaves a part of \Res{} that no choice of the two-hop training sentences of \GA{} removes.

\paragraph{Decomposition of the gradient.}
The gradient of the loss of one training token $y$ after a context $x$ with respect to the logits is $\qmodel(\cdot \mid x) - e_y$.
Here $e_y$ is the vector with 1 at $y$ and 0 elsewhere.
The gradient splits in the same way into three terms
\begin{equation}
  \qmodel - e_y = g^{\mathrm{ref}} + g^{\Res} + g^{\epsilon},
  \qquad
  \begin{aligned}
    g^{\mathrm{ref}} &= \qmodel - \pref, \\
    g^{\Res}         &= \pref - \ptrain, \\
    g^{\epsilon}     &= \ptrain - e_y .
  \end{aligned}
  \label{eq:grad}
\end{equation}
$g^{\mathrm{ref}}$ moves the model toward \pref{}.
$g^{\Res}$ moves it toward the training frequencies.
$g^{\epsilon}$ depends on which token the example shows.
It averages to zero over the training set.
Its covariance $\mathrm{diag}(\ptrain) - \ptrain \ptrain^\top$ does not depend on the model.
$g^{\epsilon}$ vanishes only where the context determines the next token.
It therefore remains after the training set is fitted.

%% file: sections/loss-results.tex
\subsection{Removing terms of the gradient}\label{sec:ladder}

Removing $g^{\epsilon}$ raises the final OOD accuracy.
Removing $g^{\Res}$ as well raises it no further.
Neither change brings the final OOD accuracy to that of answer-only training.

\paragraph{Conditions.}
We change the targets at the three undetermined positions and keep the inputs and all other targets.
The targets at all other positions are one-hot in every condition.
These include the answer and the final \tok{.}.
\Cref{tab:ladder-design} lists four conditions.
Hard targets are the targets of ordinary training.
\ptrain{} targets replace each sampled token by $\ptrain(\cdot \mid x)$ and remove $g^{\epsilon}$.
\pref{} targets replace it by $\pref(\cdot \mid x)$ and remove $g^{\Res}$ as well.
Answer-only training has no target at these positions.
Adjacent conditions differ in one term of \cref{eq:grad}.
For a given seed all conditions train on the same sentences in the same order.
\pref{} targets tell the model that a \GB{} description may continue with \tok{'s}.
They do not name the bridge or the answer.

\begin{table}[t]
\caption{The four conditions that remove terms of the gradient.}
\label{tab:ladder-design}
\centering
\small
\begin{tabular}{@{}lll@{}}
\toprule
condition & target & terms of the gradient \\
\midrule
hard targets         & $e_y$   & $g^{\mathrm{ref}} + g^{\Res} + g^{\epsilon}$ \\
\ptrain{} targets    & \ptrain & $g^{\mathrm{ref}} + g^{\Res}$ \\
\pref{} targets      & \pref   & $g^{\mathrm{ref}}$ \\
answer-only          & none    & none \\
\bottomrule
\end{tabular}
\end{table}

\begin{figure*}[t]
\centering
\includegraphics[width=\textwidth]{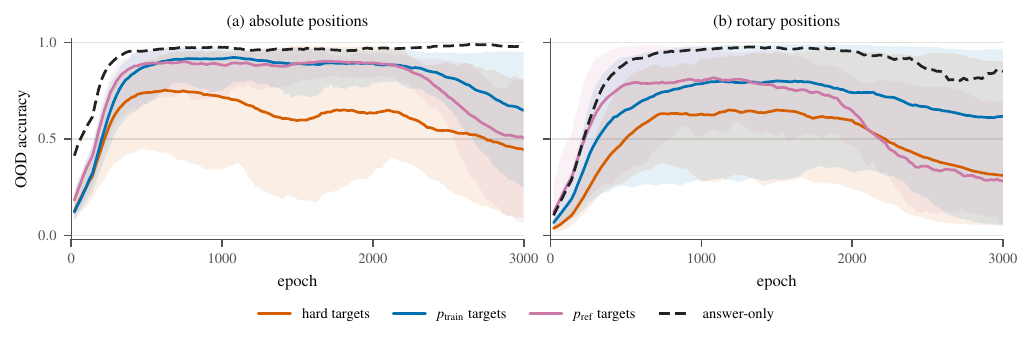}
\caption{Moving mean of the OOD accuracy of \mqal{} under the four conditions.
Lines show the mean over six seeds and bands show the range.}
\label{fig:ladder}
\end{figure*}
\begin{table*}[t]
\caption{Peak and final level of \mqal{} under the four conditions (six seeds, mean with the range in parentheses).
The last column counts the seeds whose final level is at least 0.10 below that of answer-only training.}
\label{tab:ladder}
\centering
\small
\input{tables/ladder}
\end{table*}

\paragraph{Results with \ptrain{} targets.}
With absolute positions \ptrain{} targets end at least 0.10 above hard targets in five of six seeds (\cref{fig:ladder}a and \cref{tab:ladder}).
With rotary positions four of six seeds end at least 0.10 above hard targets (\cref{fig:ladder}b).
\ptrain{} targets raise the peak (0.93 against 0.78 with absolute positions) and delay the decline.
With absolute positions most runs stay near 0.9 until about epoch 2{,}000 and decline afterwards.
\ptrain{} targets still end below answer-only training in four of six seeds with absolute positions and in three of six with rotary positions.

\paragraph{Results with \pref{} targets.}
\pref{} targets end no higher than \ptrain{} targets.
Only one of six seeds with absolute positions and none with rotary positions ends 0.10 above \ptrain{} targets.
On average \pref{} targets end lower (0.58 against 0.73 with absolute positions and 0.32 against 0.63 with rotary positions).
They end below answer-only training in all six seeds with either encoding.
The decline of OOD accuracy therefore needs neither $g^{\epsilon}$ nor $g^{\Res}$.%
\footnote{A condition between \ptrain{} and \pref{} targets could remove only the part of \Res{} that comes from the drawn two-hop questions. We did not run it, since \pref{} targets show no gain over \ptrain{} targets.}
With rotary positions answer-only training also ends at least 0.10 below its peak in four of six seeds (final levels 0.61--0.88).
The comparisons are therefore less sharp with rotary positions.
\Cref{app:optimizer} shows that neither weight decay nor the decay of the learning rate alone removes the decline.

\paragraph{Context targets.}
We call all targets of a sentence except the answer its context targets.
With \pref{} targets, removing the context targets of two-hop sentences keeps the final OOD accuracy at 0.99--1.00 in all six seeds.
Removing those of atomic sentences instead lowers it (0.41 on average against 0.58).
With \pref{} targets the decline thus comes from the context targets of two-hop sentences (\cref{app:masks}).

\subsection{Location of the decline}\label{sec:where}

We find that the first hop still writes the correct bridge into its slot.
The decline happens in loop 2 at the answer position.
Most of it is already present before the last FFN of loop 2.

\begin{table*}[t]
\caption{OOD accuracy of \mqal{} at the answer position read before the last FFN of loop 2 and as trained, at epochs 2{,}000 and 3{,}000 (absolute positions, mean over six seeds).
The last column gives the range over seeds of the share of the questions answered correctly before this FFN that it changes to a wrong answer at epoch 3{,}000.
This share is 0.00 for the ID questions in every condition and seed.}
\label{tab:a7}
\centering
\small
\input{tables/a7}
\end{table*}

\paragraph{First hop.}
During the decline the model still finds the bridge.
At epoch 3{,}000 the state that writes the bridge into its slot still predicts it for 0.86--1.00 of the OOD questions in 17 of the 18 runs with hard, \ptrain{} or \pref{} targets (0.78 in the other run).
The atomic facts of \GB{} are still answered at 1.00 in every run.
We also write the embedding of the true bridge into the slot and measure OOD accuracy again.
This raises the OOD accuracy at epoch 3{,}000 by at most 0.18.
The two accuracies decline together.

\paragraph{Second hop.}
\Cref{tab:a7} also reads the answer at the answer position just before the last FFN of loop 2.
Between epochs 2{,}000 and 3{,}000 the OOD accuracy there falls about as much as the OOD accuracy as trained.
With \pref{} targets it falls from 0.84 to 0.53 while the OOD accuracy as trained falls from 0.88 to 0.52.
With \ptrain{} targets the two accuracies fall by 0.20 and 0.26.
With hard targets both fall by 0.18.
Most of the decline thus happens before the last FFN.
The last FFN also changes some correct OOD answers to wrong ones late in training (up to 0.59 of them with \pref{} targets).
With answer-only training this almost never happens (at most 0.05).
No ID answer is changed in any condition (\cref{tab:a7}).
Rotary positions give the same picture (\cref{app:rotary}).
The decline is thus a change in the second hop at the answer position.
It affects only the OOD questions.

%% file: tables/ladder.tex
\begin{tabular}{@{}llccc@{}}
\toprule
positions & targets & peak & final level & below answer-only \\
\midrule
absolute & hard targets & 0.78\,{\scriptsize(0.50--0.98)} & 0.49\,{\scriptsize(0.14--0.86)} & 6 of 6 \\
 & $p_{\mathrm{train}}$ targets & 0.93\,{\scriptsize(0.81--0.98)} & 0.73\,{\scriptsize(0.41--0.95)} & 4 of 6 \\
 & $p_{\mathrm{ref}}$ targets & 0.93\,{\scriptsize(0.83--0.98)} & 0.58\,{\scriptsize(0.21--0.82)} & 6 of 6 \\
 & answer-only & 0.99\,{\scriptsize(0.98--1.00)} & 0.99\,{\scriptsize(0.96--1.00)} & -- \\
\midrule
rotary & hard targets & 0.69\,{\scriptsize(0.38--0.96)} & 0.35\,{\scriptsize(0.06--0.93)} & 5 of 6 \\
 & $p_{\mathrm{train}}$ targets & 0.83\,{\scriptsize(0.36--0.99)} & 0.63\,{\scriptsize(0.08--0.97)} & 3 of 6 \\
 & $p_{\mathrm{ref}}$ targets & 0.89\,{\scriptsize(0.77--0.99)} & 0.32\,{\scriptsize(0.11--0.73)} & 6 of 6 \\
 & answer-only & 0.99\,{\scriptsize(0.98--0.99)} & 0.82\,{\scriptsize(0.61--0.96)} & -- \\
\bottomrule
\end{tabular}

%% file: tables/a7.tex
\begin{tabular}{@{}lccccc@{}}
\toprule
targets & \multicolumn{2}{c}{before the last FFN} & \multicolumn{2}{c}{as trained} & changed to wrong \\
\cmidrule(lr){2-3}\cmidrule(lr){4-5}
 & 2{,}000 & 3{,}000 & 2{,}000 & 3{,}000 & 3{,}000 \\
\midrule
hard targets & 0.50 & 0.32 & 0.61 & 0.43 & 0.05--0.57 \\
$p_{\mathrm{train}}$ targets & 0.61 & 0.41 & 0.88 & 0.62 & 0.01--0.17 \\
$p_{\mathrm{ref}}$ targets & 0.84 & 0.53 & 0.88 & 0.52 & 0.05--0.59 \\
answer-only & 0.91 & 0.96 & 0.96 & 0.98 & 0.00--0.05 \\
\bottomrule
\end{tabular}

%% file: sections/related.tex
\section{Related work}\label{sec:related}

\paragraph{Latent composition of stored facts.}
Transformers can learn to compose stored facts, but only for some entities.
\citet{wang2024grokked} train a transformer on atomic facts and on two-hop compositions of some of them.
After long training the model answers two-hop questions only about entities that appear in these compositions.
The same holds for biographies written in natural language \citep{karmim2026multihop}.
\citet{ye2025implicit} find a similar limit for the second hop.
\citet{zhang2026knowing} trace such failures to upper layers that map the intermediate representation to an output instead of reasoning over it.
Pretrained language models also struggle with latent multi-hop questions \citep{yang2024latently,biran2024hopping,balesni2024lessons}.
They fail to manipulate stored knowledge without a chain of thought \citep{allenzhu2025physics}.
\citet{feng2025extractive} find that the second hop relies on the late MLP layers of a pretrained model.
In our models the decline of OOD accuracy also happens in the second hop (\cref{sec:where}).
The controlled studies train with the answer-only loss.
The studies of pretrained language models use the next-token loss but cannot separate its terms.
Our setting combines the control of the former with the loss of the latter.

\paragraph{Looped transformers.}
Looped transformers apply one stack of blocks several times \citep{dehghani2019universal}.
They can emulate programs \citep{giannou2023looped}, learn learning algorithms \citep{yang2024looped}, generalize to longer inputs \citep{fan2025looped} and reason in latent form \citep{saunshi2025reasoning}.
Recurrent depth also scales to language models with billions of parameters \citep{geiping2025scaling,zhu2025ouro}.
Under the answer-only loss looped models can answer OOD two-hop questions when two-hop facts are frequent.
\citet{wang2024grokked} report this for a model whose two halves share their weights, with 12.6 two-hop facts per atomic fact (their App.~E.2).
\citet{kohli2026loop} report it for a looped transformer with about 6.8.
An eight-layer transformer with the same compute fails there.
In our setting \bloop{} does not generalize OOD with two or four two-hop sentences per atomic fact (\cref{app:ratio}).
Several methods pass an intermediate result on without writing it out as text.
Coconut feeds the last hidden state back as the next input embedding \citep{hao2024coconut}.
Implicit chain of thought distills explicit reasoning steps into hidden states \citep{deng2023implicit}.
\bdisco{} adds the decoded prediction of every position back to that position \citep{fu2026discoloop}.
Our stream $Q$ resembles the query stream of XLNet \citep{yang2019xlnet}.
Baton passes the bridge to loop 2 as a token embedding (Condition~2).
Its answer position cannot read a decode of its own loop-1 answer (Condition~3).

\paragraph{Grokking, transient generalization and memorization.}
Small transformers can generalize long after they fit their training set \citep{power2022grokking,nanda2023progress,liu2023omnigrok}.
\citet{varma2023grokking} explain this grokking by a competition between a generalizing and a memorizing circuit.
Generalization can also appear early and fade later \citep{singh2023transient,singh2025coopetition,prakash2025grokking}.
Longer pretraining can make language models harder to fine-tune \citep{springer2025overtrained}.
The cross-entropy rewards memorization beyond what transfers \citep{achille2018emergence,harutyunyan2020improving}.
\citet{morris2026memorize} measure such memorization against a reference model.
Our residual \Res{} measures what reproducing the training frequencies gains over a reference distribution \pref{}.
The known generator lets us compute it exactly.

\paragraph{Targets and optimization.}
\citet{menon2021statistical} show that training on Bayes class probabilities instead of sampled labels lowers the variance of the empirical risk.
Our \ptrain{} targets are the conditional distribution of the training set.
They keep the empirical objective and remove only $g^{\epsilon}$.
Label noise in stochastic gradient descent acts as an implicit regularizer near the set of minimizers \citep{blanc2020implicit,haochen2021shape,damian2021label,li2022happens}.
It could thus drive late changes after the fit.
Yet the decline of OOD accuracy remains after we remove this label noise $g^{\epsilon}$.
Adaptive optimizers can also change generalization late in training \citep{thilak2022slingshot}.
\Cref{app:optimizer} tests Adam's $\epsilon$, weight decay and the decay of the learning rate.

%% file: sections/limitations.tex
\section{Limitations}\label{sec:limitations}

Our results come from one synthetic setting.
It has one sentence template, two hops and small models.
With rotary positions the comparisons are less sharp (\cref{app:rotary}).
In natural text the words before an answer vary.
A $Q$ biased toward \tok{is} may then not suffice.
The baselines may also need more two-hop sentences per atomic fact \citep[App.~E.2]{wang2024grokked}.
With twice as many they still do not generalize OOD in our setting (\cref{app:ratio}).

\Res{} and \pref{} are exact only because the generator is known.
In natural text they can only be estimated.
The \ptrain{} and \pref{} targets and the masks are diagnostics and not training methods.

The mechanism of the decline of OOD accuracy remains open.
We know where it happens (the second hop, \cref{sec:where}).
With \pref{} targets we also know which targets drive it (the context targets of two-hop sentences, \cref{sec:ladder}).
Removing $g^{\epsilon}$ and $g^{\Res}$ does not remove it.
No single setting of the optimizer removes it (\cref{app:optimizer}).
We do not know how these targets change the second hop.

%% file: sections/conclusion.tex
\section{Conclusion}\label{sec:conclusion}

We asked whether OOD generalization in two-hop composition appears under next-token training and whether it is kept.
In our setting it appears only in models that meet three design conditions.
The looped transformer, the eight-layer transformer and \bdisco{} each violate one condition.
None of them generalizes OOD under either loss.
Under next-token training \mqal{} generalizes OOD but does not keep it.
The decline of OOD accuracy remains after we remove the two terms of the gradient that depend on the training set.
With \pref{} targets, removing the context targets of two-hop sentences removes the decline.
During the decline the model still recalls both facts.
It loses the answer in the second hop.

In our setting the context targets of two-hop sentences lower OOD accuracy even when they equal \pref{}.
Language-model corpora contain many sentences that compose facts.
Whether their context targets also limit latent composition in language models is open.
The three conditions also need a form that does not rely on our sentence format.

%% file: sections/appendix.tex
\input{sections/appendix/config}
\FloatBarrier
\input{sections/appendix/duplicate}
\FloatBarrier
\input{sections/appendix/residual}
\FloatBarrier
\input{sections/appendix/optimizer}
\FloatBarrier
\input{sections/appendix/masks}
\FloatBarrier
\input{sections/appendix/ratio}
\FloatBarrier
\input{sections/appendix/rotary}
\FloatBarrier
\input{sections/appendix/seeds}
\FloatBarrier

%% file: sections/appendix/config.tex
\section{Configurations and training details}\label[appendix]{app:config}

This appendix gives the details of the models, of training and of the training sets.
It also reports how \bdisco{} behaves in our settings.

\subsection{Models}\label{app:models}

\paragraph{Backbone.}
Every model is built from the same transformer block.
A block applies a LayerNorm, causal self-attention (4 heads of width 64) and a residual connection.
It then applies a second LayerNorm, a feed-forward layer of width 1{,}024 with a GELU activation and a second residual connection.
All linear layers have biases.
A learned position vector $p_i$ is added to the embedding of the token at position $i$.
\Cref{app:rotary} replaces these vectors by rotary position embeddings with base $10^4$ in every attention layer.
The vocabulary has 1{,}053 tokens (1{,}000 entities, 50 relations and the three function words) and a padding token.
The logits are computed from a state by a final LayerNorm followed by the tied embedding matrix $W$.
The embeddings and the position vectors are initialized from $\mathcal{N}(0, 0.02^2)$.
The linear layers use the defaults of PyTorch.
The vector $\delta$ starts at zero.

\paragraph{Variants and baselines.}
\Cref{tab:models} lists the models.
The looped models apply the same 4 blocks in loop 1 and in loop 2.
Their loss is computed from the logits after loop 2.
They differ only in what happens between the two loops.
\Cref{sec:design} gives what each stream of the three variants reads (\cref{eq:readH,eq:readQ}).
We add four details here.
The slot $C_i$ receives its own learned position vector $s_i$ in both loops.
In \mbase{} the slots pass through no layer.
In every block the attention of $H$ reads keys and values that the block computes from the slots with its own projections.
In \mq{} and \mqal{} a slot attends only to itself.
The output of its attention is therefore its own value.
The slot then passes the feed-forward layer.
The input $W_{\mathtt{is}} + \delta$ of $Q_i$ is the same vector at every position.
$Q_i$ receives $p_{i+1}$ in \mq{}.
\mq{} therefore has one more position vector for the $Q$ at the last position.

\begin{table}[htbp]
\caption{The six models with absolute positions.
$\phi$ is the decode of \cref{eq:phi}.
\emph{Added} gives the parameters that a model has in addition to those of \bloop{}.}
\label{tab:models}
\centering
\small
\begin{tabular}{@{}llllr@{}}
\toprule
model & blocks & between loop 1 and loop 2 & added & parameters \\
\midrule
\btf{}    & 8 distinct, no loop & --                                                              & --                                        & 6.59M \\
\bloop{}  & 4, applied twice    & nothing                                                         & --                                        & 3.43M \\
\bdisco{} & 4, applied twice    & $H_i \leftarrow H_i + \mathrm{RMSNorm}(\phi(H_i))$ at every $i$ & 256 (RMSNorm gain)                        & 3.43M \\
\mbase{}  & 4, applied twice    & $C_i \leftarrow \phi(H_i)$                                      & 2{,}048 ($s_i$)                           & 3.43M \\
\mqal{}   & 4, applied twice    & $C_i \leftarrow \phi(Q_i)$                                      & 2{,}304 ($s_i$ and $\delta$)              & 3.43M \\
\mq{}     & 4, applied twice    & $C_i \leftarrow \phi(Q_i)$                                      & 2{,}560 ($s_i$, $\delta$ and $p_{T+1}$)   & 3.43M \\
\bottomrule
\end{tabular}
\end{table}

\subsection{Training and evaluation}\label{app:training}

\paragraph{Optimizer and schedule.}
Every run uses AdamW with learning rate $10^{-3}$, $\beta = (0.9, 0.98)$, $\epsilon = 10^{-8}$ and weight decay 0.1 on all parameters.
\Cref{app:optimizer} changes $\epsilon$, the weight decay or the schedule in a few runs.
The learning rate rises linearly over the first 1{,}000 steps.
It then follows a cosine decay to 0.1 of its peak at the last step.
Gradients are clipped to a norm of 1.0.
The forward pass runs in bfloat16.
A batch holds 512 sentences.
An epoch is one pass over the training set in a random order.
Every run trains for 3{,}000 epochs.
With 20{,}000 training sentences an epoch has 40 steps and a run has 120{,}000 steps.

\paragraph{Random seeds.}
The world, the training sets and the test questions are generated once and are the same in every run.
The seed of a run sets the initialization and the order of the batches.
\Cref{tab:configs} lists the configurations and their numbers of seeds.

\paragraph{Evaluation.}
Every 20 epochs we measure the accuracy at the answer position on the atomic facts, on the two-hop training sentences, on the 2{,}000 ID questions and on the 2{,}000 OOD questions.
The curves of \cref{fig:ood,fig:ladder,fig:masks} show the moving mean of each run over 13 measurements (250 epochs) centered on each point.
The window is shorter at the start and at the end of training.
The peak of a run is the highest moving mean over full windows.
The final level is the mean over the last 25 measurements (epochs 2{,}520--3{,}000).

\paragraph{Snapshots and selected accuracy.}
Most runs keep a snapshot of the model every 100 epochs.
The OOD questions are split once at random into a validation half and a test half of 1{,}000 questions each.
The selected OOD accuracy is the accuracy on the test half at the snapshot with the highest accuracy on the validation half.
It needs OOD questions for the choice of the snapshot.
We therefore report it only with the results of each seed (\cref{app:seeds}).
The branches of \cref{app:optimizer} also start from a snapshot.

\begin{table}[htbp]
\caption{The configurations and their numbers of seeds.
All runs use the settings of \cref{app:training} and absolute positions unless the row says otherwise.}
\label{tab:configs}
\centering
\small
\setlength{\tabcolsep}{4pt}
\begin{tabular}{@{}llll@{}}
\toprule
section & models & loss and targets & seeds \\
\midrule
\cref{sec:arch-results} & \mbase{}, \mq{}                       & answer-only, next-token & 4 \\
                        & \mqal{}                               & answer-only, next-token & 6 \\
                        & \bloop{}, \btf{}, \bdisco{}           & answer-only, next-token & 4 \\
\midrule
\cref{sec:tests}        & \mbase{} without the target of $H_4$  & next-token              & 4 \\
                        & \mq{} with the duplicate slot hidden  & answer-only, next-token & 4 \\
                        & \mqal{} with an undecoded slot        & answer-only, next-token & 4 \\
\midrule
\cref{sec:ladder}       & \mqal{}                               & hard, \ptrain{} and \pref{} targets, answer-only & 6 \\
                        & \mqal{}, rotary positions             & hard, \ptrain{} and \pref{} targets, answer-only & 6 \\
\midrule
\cref{app:masks}        & \mqal{}                               & five conditions with targets removed & 6 \\
                        & \mqal{}, rotary positions             & \pref{} targets without the $r_2$ target & 4 \\
\cref{app:optimizer}    & \mqal{}                               & \pref{} targets, four branches at epoch 2{,}000 & 4 \\
\cref{app:ratio}          & \bloop{}, \btf{}                      & answer-only, next-token, 20{,}000 two-hop sentences & 4 \\
                        & \mqal{}                               & next-token, 20{,}000 two-hop sentences & 4 \\
                        & \bloop{}, \btf{}, rotary positions    & next-token, 20{,}000 two-hop sentences & 4 \\
\cref{app:rotary}       & \mbase{}, \mq{}, \mqal{}, rotary positions & answer-only, next-token & 4--6 \\
\cref{app:disco}        & \bloop{}, \btf{}, \bdisco{}, rotary positions & answer-only, next-token & 4--6 \\
                        & \bdisco{} and baselines, symbolic format & see \cref{tab:disco,tab:disco-variants} & 1--2 \\
\bottomrule
\end{tabular}
\end{table}

\subsection{Training sets and changes for the tests}\label{app:data}

\paragraph{Training sets.}
The 50{,}000 chains of \GA{} are shuffled once.
The first 10{,}000 are the two-hop training sentences.
The next 2{,}000 are the ID questions.
The OOD questions are 2{,}000 chains of \GB{} drawn at random.
\Cref{app:ratio} uses a training set with 20{,}000 random chains of \GA{} instead.

\paragraph{Changes for the tests and analyses.}
\Cref{sec:hypothesis,sec:nexttoken} change the targets or the model in a few ways.
\ptrain{} targets replace the observed token at every position by the frequencies of the next token after its context in the full training set.
They differ from the observed token only at the three contexts of \cref{tab:pref}.
\pref{} targets use the \pref{} of \cref{app:residual} at the same three contexts.
A removed target contributes zero to the average over all targets in \cref{eq:loss} (\cref{app:masks}).
The duplicate slot is hidden only in loop 2.
In loop 1 the slot holds the input token as in every other run.
The undecoded slot holds the state $Q_i$ after the final LayerNorm.
This state is rescaled to the mean norm of the token embeddings.
The gradient does not flow through this norm.
A branch loads the model, the optimizer state and the random state of a run at epoch 2{,}000.
It then continues with one setting changed (\cref{app:optimizer}).

\subsection{\bdisco{} in our settings}\label{app:disco}

\citet{fu2026discoloop} introduced the decode $\phi$ that the slots of our variants receive.
This section reports how \bdisco{} behaves in our settings.

\paragraph{Implementation of \bdisco{}.}
We implement \bdisco{} as defined by \citet[Eqs.~4--5]{fu2026discoloop}, with the fixed gate $\alpha = 1$ and the temperature $\tau = 1$ of their symbolic runs.
Between loop 1 and loop 2, every position $i$ receives $H_i \leftarrow H_i + \mathrm{RMSNorm}(\phi(H_i))$, where $\phi$ is the decode of \cref{eq:phi} and the RMSNorm has a learned gain.
Our $\phi$ reads the logits after the final LayerNorm, as our output layer does.
\citet{fu2026discoloop} write these logits as $Wh$.
We also ran this form (\cref{tab:disco-variants}).
Their paper does not specify the size of the model or the training settings of the synthetic task, so we use those of \cref{app:models,app:training}.

\paragraph{Symbolic format and loop-1 accuracy.}
Besides our main setting, we train in the symbolic format of \citet{fu2026discoloop}.
In this format an atomic fact is the three tokens $a$~$r$~$b$.
A two-hop question is the four tokens $a$~$r_1$~$r_2$~$c$.
The loss is on the last token.
These runs add learned absolute position vectors to the input.
They keep the learning rate constant after the warm-up unless we state otherwise.
The rest is as in \cref{app:training}.
To see how a model answers we also read its logits after loop 1.
We call the resulting accuracy the loop-1 accuracy.

\paragraph{Re-entry at the bridge and at the answer position.}
The re-entry of \bdisco{} acts at every position.
Two positions matter for a two-hop question.
At the position of $r_1$, the decode carries the bridge, which loop 2 needs for the second hop.
At the answer position, the decode carries the loop-1 prediction of the answer itself.
When loop 1 already predicts the answer there, loop 2 can copy it.
We study the two parts separately.

\paragraph{Training without re-entry at the answer position.}
We first train \bdisco{} in the symbolic format without the re-entry at the answer position.
We keep the re-entry at every other position.
This change needs to know the answer position, so it is a diagnostic and not an architecture for language models.
With a cosine schedule, its two seeds reach an ID accuracy of 1.00 and OOD accuracies of 0.97 and 1.00.
These are the accuracies that \citet{fu2026discoloop} report for \bdisco{} (\cref{tab:disco}).
The loop-1 accuracy on the two-hop training sentences falls below 0.1 by epoch 180.
\citet{fu2026discoloop} observe the same fall near epoch 500.
We then check that these models answer through the bridge.
Between the loops, we replace the re-entry term at the position of $r_1$ by that of another entity $b'$ that has the relation $r_2$.
The final answer then becomes the object of $(b', r_2)$ for 0.94 and 0.99 of the OOD questions.
This agrees with the intervention of \citet[Sec.~3.2]{fu2026discoloop}, which places the embedding of the decoded bridge at the position of $r_1$ between the loops.
With a constant learning rate, the highest OOD accuracies during training are 0.98 and 0.96, but OOD accuracy varies widely between evaluations and ends at 0.43 and 0.13.
With rotary positions and a cosine schedule, the variant reaches ID accuracies of 0.99 and 1.00 and OOD accuracies of 0.40 and 0.69.

\paragraph{Re-entry at every position.}
\citet{fu2026discoloop} report that \bdisco{} generalizes OOD on their symbolic task, with ID and OOD accuracies close to 1.00.
They report a similar result on a synthetic-language task whose direct questions, such as ``Finley's teacher's wife is Charlie'', are close to our mini-language.
In our settings, \bdisco{} with the re-entry at every position does not generalize OOD (\cref{tab:disco}).
In the symbolic format, its ID accuracy is 0.11 and its OOD accuracy 0.00.
In our mini-language its OOD accuracy stays at or below 0.01 at every evaluation under either loss and with either position encoding.
We ran it 32 times, in the two formats and with the changes of \cref{tab:disco-variants}.
In each run, the loop-1 accuracy on the two-hop training sentences reaches 0.99 between epochs 120 and 480.
It stays above 0.93 from epoch 200 on and ends at 0.98--1.00.
These models thus answer the two-hop training sentences in loop 1.

\begin{table}[htbp]
\caption{ID and OOD accuracy at the end of training, as reported by \citet{fu2026discoloop} and as measured by us at the last evaluation (epoch 3{,}000).
\emph{Symbolic} is their format, in which the loss is on the last token.
\emph{Mini-language} is our format (\cref{sec:setting}).
The non-looped model stacks two copies of the blocks of \bloop{} with distinct weights.
Ours is \btf{}.
Their values marked $\approx$ are read from their Figure~3.
Their other values are stated in their text and Table~1.
Our values are means over seeds on all 2{,}000 questions.}
\label{tab:disco}
\centering
\small
\begin{tabular}{@{}lllcccc@{}}
\toprule
 & & & & non-looped & \bloop{} & \bdisco{} \\
source & format & positions & seeds & ID / OOD & ID / OOD & ID / OOD \\
\midrule
\citet{fu2026discoloop} & symbolic & not reported & not reported & $<$0.20 / $\approx$0.00 & 0.71 / 0.08 & $\approx$1.00 / $\approx$0.98 \\
ours & symbolic & absolute & 2 & 0.20 / 0.00 & 0.78 / 0.00 & 0.11 / 0.00 \\
ours & symbolic & rotary   & 2 & 0.14 / 0.00 & 0.13 / 0.00 & 0.09 / 0.00 \\
ours & mini-language, answer-only & rotary & 4   & 0.06 / 0.00 & 0.05 / 0.00 & 0.04 / 0.00 \\
ours & mini-language, next-token  & rotary & 4--6 & 0.05 / 0.00 & 0.04 / 0.00 & 0.05 / 0.00 \\
ours & mini-language, answer-only & absolute & 4 & 0.09 / 0.00 & 0.15 / 0.00 & 0.05 / 0.00 \\
ours & mini-language, next-token  & absolute & 4 & 0.04 / 0.00 & 0.03 / 0.00 & 0.04 / 0.00 \\
\bottomrule
\end{tabular}
\end{table}

\paragraph{Substitution of the re-entry terms.}
The final answer of these models follows the re-entry at the answer position.
We test this without further training, on the two runs in the symbolic format.
Between the loops, we replace the re-entry term at the answer position by that of a wrong entity of the same graph, at the same norm.
The final answer then becomes this entity for 0.99 of the two-hop training sentences and for all ID and OOD questions.
When we instead replace the term at the position of $r_1$ by that of $b'$, at most 0.01 of the final answers become the object of $(b', r_2)$.
Loop 2 thus copies the loop-1 answer and does not use the bridge.
The rule of \cref{sec:design} excludes this copy, since $H_i$ reads only the slots of earlier positions.

\paragraph{Variants of \bdisco{}.}
In the symbolic format with absolute positions, we also changed one setting of \bdisco{} at a time (\cref{tab:disco-variants}).
The changes include their learnable gate (their Eq.~6), a weaker re-entry, other learning rates and schedules, and a larger weight decay.
None of these runs generalizes OOD.
Their final OOD accuracy is at most 0.002.
Their OOD accuracy never exceeds 0.02 during training.

\begin{table}[htbp]
\caption{\bdisco{} with one setting changed, in the symbolic format with absolute positions.
Final ID and OOD accuracy with one or two seeds per row.
Rows with two seeds give both values.
The rows with a cosine schedule decay the learning rate as in \cref{app:training}.}
\label{tab:disco-variants}
\centering
\small
\begin{tabular}{@{}lccc@{}}
\toprule
change & seeds & ID & OOD \\
\midrule
none (\cref{tab:disco})                           & 2 & 0.12 / 0.10 & 0.00 \\
learnable gate (their Eq.~6)                      & 2 & 0.12 / 0.12 & 0.00 \\
RMSNorm without gain                              & 1 & 0.13        & 0.00 \\
cosine schedule                                   & 1 & 0.10        & 0.00 \\
learning rate $10^{-4}$                           & 1 & 0.05        & 0.00 \\
$\alpha = 0.1$, no gain, cosine schedule          & 2 & 0.24 / 0.22 & 0.00 \\
$\alpha = 0.3$, no gain, cosine schedule          & 2 & 0.13 / 0.11 & 0.00 \\
weight decay 0.3, cosine schedule                 & 1 & 0.10        & 0.00 \\
weight decay 1.0, cosine schedule                 & 1 & 0.07        & 0.00 \\
logits $Wh$ inside $\phi$, cosine schedule        & 1 & 0.17        & 0.00 \\
\bottomrule
\end{tabular}
\end{table}

\paragraph{Baselines and position encoding.}
In the symbolic format with absolute positions, our non-looped model and our \bloop{} come close to the accuracies that \citet{fu2026discoloop} report (ID 0.20 and 0.78, against below 0.20 and 0.71).
Our \bloop{} also answers the two-hop training sentences in loop 2.
Its loop-1 accuracy on the two-hop training sentences ends at 0.00, while its final accuracy on them is 1.00.
Rotary positions change \bloop{}.
With them \bloop{} answers the two-hop training sentences in loop 1 (loop-1 accuracy 0.99).
Its ID accuracy falls from 0.78 to 0.13.
The position encoding thus changes \bloop{} strongly in the symbolic format.
In our mini-language the mean ID accuracy of \bloop{} at the last evaluation is at most 0.15 with either encoding (\cref{tab:disco}).
Its OOD accuracy is at most 0.01 in every setting.

\paragraph{Summary.}
In our settings, the re-entry of the bridge lets \bdisco{} generalize OOD with the accuracies that \citet{fu2026discoloop} report.
The re-entry at the answer position lets loop 2 copy the answer of loop 1.
With it \bdisco{} does not generalize OOD.
We do not know which of the settings that their paper leaves open lets their runs avoid this copy.

%% file: sections/appendix/duplicate.tex
\section{The duplicate slot}\label[appendix]{app:duplicate}

\Cref{sec:conditions} describes the duplicate slot of \mq{}.
This appendix measures how the atomic sentences of \GB{} use it.

\begin{table}[htbp]
\caption{The duplicate slot on atomic facts of \GB{} at epoch 500 and at the end of training (absolute positions, four seeds, mean with the range in parentheses).
\emph{Follow} is the share of answers that change to the object of another fact written into the duplicate slot after loop 1.
\emph{Margin} is the median margin of $H_4$ after loop 1.
\mqal{} has no duplicate slot.}
\label{tab:duplicate}
\centering
\small
\input{tables/duplicate}
\end{table}

\paragraph{Measurements.}
We measure two quantities on atomic facts of \GB{} (\cref{tab:duplicate}).
The first is the share of answers that follow the duplicate slot.
After loop 1 we write the object of another fact into the duplicate slot $C_3$.
We count the answers that change to this object.
The second is how strongly $H_4$ at \tok{is} ranks the correct answer after loop 1.
We take the logit of the correct answer minus the largest logit of any other token.
We call this difference the margin.
A positive margin means that the computation of $H_4$ itself ranks the correct answer first.
We report the median of the margin over the atomic facts that the OOD questions use as their second hop.

\paragraph{The answer-only loss.}
Under the answer-only loss the atomic answers of \GB{} move to the duplicate slot during training.
At epoch 500 between 0.24 and 0.43 of them follow it.
At the end 0.98--1.00 do.
Over the same epochs the margin of $H_4$ for \GB{} falls from 4.1--4.3 to between $-1.6$ and 4.2.
The atomic answers of \GA{} stay with $H_4$.
At the end none of them follow the duplicate slot.
Their margin grows from 7.0 to 13.5 on average.
The two-hop sentences of \GA{} keep training the computation of Condition~1 for \GA{} bridges.
For \GB{} entities only atomic sentences train this computation.
These sentences now use the duplicate slot.
The second hop of an OOD question therefore runs a computation that has weakened.
This matches the course of \mq{} under the answer-only loss (peak 0.78--0.91 and final level 0.06--0.56, \cref{tab:ood}).
Without a duplicate slot the margin of \mqal{} for \GB{} stays between 6.8 and 14.6 (\cref{tab:duplicate}).
Its OOD accuracy stays as well (final level 0.99 on average, \cref{tab:ood}).

\paragraph{The next-token loss.}
\mq{} uses the duplicate slot under the next-token loss as well.
At the end every atomic answer of \GB{} follows it.
The margin of $H_4$ for \GB{} falls to 1.3--2.6.
Hiding the duplicate slot in loop 2 (\cref{sec:tests}) raises the mean final level of \mq{} under the next-token loss from 0.12 to 0.43.
Per seed it changes from 0.23, 0.02, 0.11 and 0.12 to 0.41, 0.03, 0.87 and 0.41.
In the same four seeds \mqal{} ends at 0.52 on average (0.41, 0.42, 0.39 and 0.86).
Over all six seeds of \cref{tab:ood} its mean is 0.49.

%% file: tables/duplicate.tex
\begin{tabular}{@{}llcccc@{}}
\toprule
model & loss & \multicolumn{2}{c}{follow} & \multicolumn{2}{c}{margin} \\
\cmidrule(lr){3-4}\cmidrule(l){5-6}
 & & 500 & end & 500 & end \\
\midrule
\mq{} & answer-only & 0.32\,{\scriptsize(0.24--0.43)} & 0.99\,{\scriptsize(0.98--1.00)} & 4.2\,{\scriptsize(4.1 to 4.3)} & 0.9\,{\scriptsize($-$1.6 to 4.2)} \\
\mq{} & next-token & 0.68\,{\scriptsize(0.55--0.83)} & 1.00\,{\scriptsize(1.00--1.00)} & 3.7\,{\scriptsize(3.5 to 3.9)} & 2.1\,{\scriptsize(1.3 to 2.6)} \\
\mqal{} & answer-only & -- & -- & 10.2\,{\scriptsize(6.8 to 14.6)} & 9.4\,{\scriptsize(7.3 to 11.7)} \\
\mqal{} & next-token & -- & -- & 5.6\,{\scriptsize(5.1 to 6.0)} & 6.3\,{\scriptsize(5.9 to 6.9)} \\
\bottomrule
\end{tabular}

%% file: sections/appendix/residual.tex
\section{The residual and \texorpdfstring{\pref{}}{p\_ref}}\label[appendix]{app:residual}

The residual of \cref{eq:loss} is $\Res = \E_x \KL\big(\ptrain(\cdot \mid x) \,\|\, \pref(\cdot \mid x)\big)$.
Our training sets come from a known generator.
Both distributions are therefore known at every context.
\Res{} can thus be computed exactly.
This appendix states the class of predictors that defines \pref{}, gives \pref{} and \ptrain{} at each context and computes \Res{}.

\paragraph{Contexts with an undetermined next token.}
The next-token loss predicts every token after the first.
At most contexts the atomic facts and the format of the sentences determine the next token.
For example, \tok{'s} always follows the first entity.
The object $b$ always follows $a$~\tok{'s}~$r$~\tok{is}.
At these contexts \ptrain{} and \pref{} both put all their probability on this token.
Their KL is zero.
Three contexts remain (\cref{tab:pref}).
After $a$~\tok{'s} the next token is a relation of $a$.
After $a$~\tok{'s}~$r_1$ it is \tok{is} in an atomic sentence and \tok{'s} in a two-hop sentence.
After $a$~\tok{'s}~$r_1$~\tok{'s} it is a relation $r_2$ of the bridge $b$.
These are the three contexts at which $g^{\epsilon}$ is not zero (\cref{sec:decomposition}).

\paragraph{The class of predictors.}
\pref{} is the maximum-likelihood fit to the training set within one class of predictors.
A predictor in this class treats all entities alike.
Its input is the type of the context and the facts of the two hops.
These facts are the relations of $a$, the bridge $b$ and the relations of $b$.
The class has no input for the identity of an entity or for its graph.
The graph must be excluded from the input explicitly.
The training set contains the atomic facts of both graphs.
Since the two graphs are disjoint, these facts reveal the graph of every entity.
A predictor that used them freely could therefore tell \GA{} from \GB{}.
We also assume that every relation of an entity is equally likely to be asked about.
\pref{} then follows from the training set alone.

\begin{table}[htbp]
\caption{\ptrain{} and \pref{} at the three contexts whose next token the world does not determine.
$b$ is the bridge of $(a, r_1)$.
$n(a, r)$ is the number of two-hop training sentences that begin with $a$~\tok{'s}~$r$.
In the second row $n = n(a, r_1)$.
The value $1/2$ holds for 10{,}000 two-hop sentences.}
\label{tab:pref}
\centering
\small
\begin{tabular}{@{}llll@{}}
\toprule
context & next token & \ptrain{} & \pref{} \\
\midrule
$a$~\tok{'s}                & $r$      & $\propto 1 + n(a, r)$ & uniform over the relations of $a$ \\
$a$~\tok{'s}~$r_1$          & \tok{'s} & $n / (n + 1)$ (\GA{}), 0 (\GB{}) & $1/2$ \\
$a$~\tok{'s}~$r_1$~\tok{'s} & $r_2$    & frequencies in training & uniform over the relations of $b$ \\
\bottomrule
\end{tabular}
\end{table}

\paragraph{\texorpdfstring{\pref{}}{p\_ref} at the three contexts.}
\Cref{tab:pref} gives \ptrain{} and \pref{} at the three contexts.
\pref{} is uniform over the ten relations of $a$ after $a$~\tok{'s}.
It gives \tok{'s} the same probability after the description $a$~\tok{'s}~$r_1$ of every entity.
This probability is the share of descriptions continued by \tok{'s} in the training set.
It is $1/2$ with 10{,}000 two-hop and 10{,}000 atomic sentences.
It is $2/3$ with 20{,}000 two-hop sentences (\cref{app:ratio}).
\pref{} is uniform over the ten relations of $b$ after $a$~\tok{'s}~$r_1$~\tok{'s}.
\pref{} gives a positive probability to every continuation that occurs in training.
\Res{} is therefore finite.
The \pref{} targets of \cref{sec:ladder} use \pref{} at these three contexts and the observed token at every other position.

\paragraph{Normalization per two-hop sentence.}
\Cref{eq:loss} averages over the training targets.
We report \Res{} per two-hop sentence instead, since it depends on which two-hop sentences the training set contains.
It is the sum of the KL over all training contexts divided by the number of two-hop training sentences.
With 10{,}000 two-hop sentences the training set has 120{,}000 targets (five per atomic sentence and seven per two-hop sentence).
Per target \Res{} is then 0.17 nats.
The conditional entropy \Htrain{} is 0.52 nats per target.

\begin{table*}[t]
\caption{The residual of the random training sets in nats per two-hop sentence, split over the contexts of \cref{tab:pref}.
The four columns of contexts sum to \Res{}.
\Htrain{} is the conditional entropy of the training set per target and after $a$~\tok{'s}~$r_1$~\tok{'s} alone.
The training set with 20{,}000 two-hop sentences is that of \cref{app:ratio}.}
\label{tab:residual}
\centering
\small
\begin{tabular}{@{}rccccccc@{}}
\toprule
 & \multicolumn{4}{c}{residual after} & & \multicolumn{2}{c}{\Htrain{}} \\
\cmidrule(lr){2-5}\cmidrule(l){7-8}
two-hop sentences & $a$~\tok{'s} & $a$~\tok{'s}~$r_1$ (\GA{}) & $a$~\tok{'s}~$r_1$ (\GB{}) & $a$~\tok{'s}~$r_1$~\tok{'s} & \Res{} & per target & after $a$~\tok{'s}~$r_1$~\tok{'s} \\
\midrule
10{,}000 & 0.12 & 0.19 & 0.35 & 1.37 & 2.03 & 0.52 & 0.93 \\
20{,}000 & 0.06 & 0.07 & 0.28 & 0.84 & 1.24 & 0.58 & 1.47 \\
\bottomrule
\end{tabular}
\end{table*}

\paragraph{Residual by context.}
\Cref{tab:residual} splits \Res{} over the contexts.
With 10{,}000 random two-hop sentences most of \Res{} lies after $a$~\tok{'s}~$r_1$~\tok{'s} (1.37 of 2.03).
In training a prefix of \GA{} is followed by about two of the ten relations of its bridge.
Which two is arbitrary.
\pref{} cannot predict them and spreads its probability over all ten.
Take a prefix that is followed by two different relations once each.
\ptrain{} gives each of them $1/2$ and \pref{} gives each $1/10$.
The KL is then $\log 5 = 1.61$ nats per sentence of this prefix.
Only memorizing which relations followed each prefix removes this term.
With 20{,}000 two-hop sentences a prefix is followed by about four relations.
This term then falls to 0.84.

The second largest term (0.35) comes from the descriptions $a$~\tok{'s}~$r_1$ of \GB{} entities.
In training they are always followed by \tok{is}, since no \GB{} entity begins a two-hop sentence.
\pref{} gives \tok{'s} the probability $1/2$ after every description.
Each of the 5{,}000 atomic sentences of \GB{} thus adds $\log 2$.
Per two-hop sentence this is $5{,}000 \cdot \log 2 / 10{,}000 = 0.35$.
A reference that could tell the two graphs apart would remove this term.
Its residual is 1.60.

The terms after $a$~\tok{'s} (0.12) and after the descriptions of \GA{} entities (0.19) come from the number $n(a, r)$ of two-hop sentences that begin with each prefix.
After $a$~\tok{'s} this number enters \ptrain{} through $1 + n(a, r)$.
After the description of a \GA{} entity it enters the share of \tok{'s} through $n / (n + 1)$.

%% file: sections/appendix/optimizer.tex
\section{Optimizer settings and the late decline}\label[appendix]{app:optimizer}

The decline of OOD accuracy happens late in training (\cref{sec:ladder}).
At that time the learning rate is small.
Weight decay has also acted for a long time.
This appendix tests whether a single setting of the optimizer causes the decline.

\paragraph{Branches.}
We branch the runs of \mqal{} with \pref{} targets at epoch 2{,}000 (seeds 0--3).
Each branch loads the model, the optimizer state and the random state of its run at this epoch (\cref{app:data}).
It then trains to epoch 3{,}000 with one setting changed.
The first branch changes nothing and serves as the reference.
The second sets the weight decay to 0.
The third keeps the learning rate at its value at epoch 2{,}000.
The fourth raises Adam's $\epsilon$ from $10^{-8}$ to $10^{-5}$.
Adam divides each gradient by its own recent typical size.
It therefore keeps its step size when the gradients shrink after the fit.
A larger $\epsilon$ makes the steps smaller when the gradients are small.

\begin{table}[htbp]
\caption{Final level of the OOD accuracy of \mqal{} with \pref{} targets in the four branches from epoch 2{,}000 (absolute positions).
Differences to the reference branch of the same seed are in parentheses.}
\label{tab:optimizer}
\centering
\small
\begin{tabular}{@{}ccccc@{}}
\toprule
seed & reference & weight decay 0 & learning rate frozen & $\epsilon = 10^{-5}$ \\
\midrule
0 & 0.73 & 0.93 ($+$0.20) & 0.88 ($+$0.15) & 0.83 ($+$0.10) \\
1 & 0.63 & 0.78 ($+$0.15) & 0.78 ($+$0.15) & 0.76 ($+$0.13) \\
2 & 0.82 & 0.87 ($+$0.05) & 0.80 ($-$0.03) & 0.79 ($-$0.03) \\
3 & 0.70 & 0.42 ($-$0.28) & 0.67 ($-$0.03) & 0.89 ($+$0.20) \\
\bottomrule
\end{tabular}
\end{table}

\paragraph{Results.}
The reference branch reproduces each run exactly.
Training from a snapshot is thus deterministic.
The other branches differ from the reference only through the changed setting.
No branch removes the decline (\cref{tab:optimizer}).
Weight decay 0 and the frozen learning rate each raise the final level by at least 0.10 in two of four seeds.
Weight decay 0 lowers it by 0.28 in seed 3.
With the frozen learning rate the OOD accuracy declines more slowly but still declines.
In seed 3 its moving mean falls from 0.89 at epoch 2{,}000 to 0.53 at epoch 3{,}000.
The larger $\epsilon$ raises the final level by at least 0.10 in three of four seeds.
In seed 2 the OOD accuracy still declines with it.
Its moving mean falls from 0.95 at epoch 2{,}000 to 0.68 at epoch 3{,}000.
No single setting of the optimizer is therefore the cause of the decline.

%% file: sections/appendix/masks.tex
\section{Removing context targets}\label[appendix]{app:masks}

This appendix removes the targets at some positions, starting from \pref{} targets (\cref{sec:ladder}).
In a two-hop sentence the position after the description has the input \tok{'s} and the target $r_2$ (\cref{tab:format}).
We call this target the $r_2$ target.

\begin{table*}[htbp]
\caption{Targets kept (\checkmark) and removed (--) in each condition.
Context targets are all targets except the answer.}
\label{tab:mask-design}
\centering
\small
\begin{tabular}{@{}lccccc@{}}
\toprule
 & \multicolumn{2}{c}{atomic sentence} & \multicolumn{3}{c}{two-hop sentence} \\
\cmidrule(lr){2-3}\cmidrule(lr){4-6}
condition & context & answer & other context & $r_2$ & answer \\
\midrule
\pref{} targets                          & \checkmark & \checkmark & \checkmark & \checkmark & \checkmark \\
without two-hop context targets          & \checkmark & \checkmark & --         & --         & \checkmark \\
without atomic context targets           & --         & \checkmark & \checkmark & \checkmark & \checkmark \\
without the $r_2$ target                 & \checkmark & \checkmark & \checkmark & --         & \checkmark \\
answer-only plus the $r_2$ target        & --         & \checkmark & --         & \checkmark & \checkmark \\
answer-only                              & --         & \checkmark & --         & --         & \checkmark \\
\bottomrule
\end{tabular}
\end{table*}

\begin{figure*}[htbp]
\centering
\includegraphics[width=\textwidth]{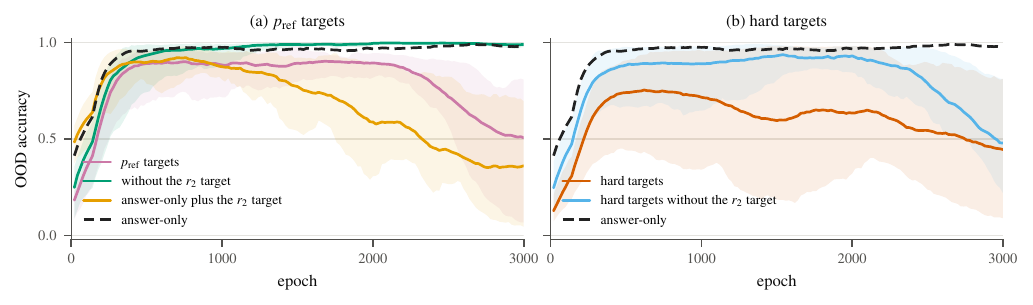}
\caption{Moving mean of the OOD accuracy of \mqal{} with targets removed (absolute positions).
Lines show the mean over six seeds and bands show the range.}
\label{fig:masks}
\end{figure*}
\begin{table*}[htbp]
\caption{Peak and final level of \mqal{} with targets removed from \pref{} targets and from hard targets (absolute positions, six seeds).
The last column counts the seeds whose final level is at least 0.10 below answer-only training.}
\label{tab:masks}
\centering
\small
\input{tables/masks}
\end{table*}
\begin{table*}[htbp]
\caption{OOD accuracy before the last FFN of loop 2 and as trained, and the share that this FFN changes from correct to wrong, for the conditions of this section (\mqal{}, absolute positions, as in \cref{tab:a7}).}
\label{tab:a7-masks}
\centering
\small
\input{tables/masks_a7}
\end{table*}

\paragraph{Conditions.}
\Cref{tab:mask-design} lists the conditions.
They start from \pref{} targets and remove the targets at some positions or keep only some.
A removed target contributes zero to the average over all targets in \cref{eq:loss}.
The remaining targets thus keep their weight.

\paragraph{Context targets.}
Removing the context targets of two-hop sentences keeps the final OOD accuracy at 0.99--1.00 in all six seeds, as in answer-only training (\cref{tab:masks}).
Atomic sentences keep all their targets in this condition.
Removing the context targets of atomic sentences instead lowers the final OOD accuracy (0.41 against 0.58 with all \pref{} targets).
The context targets of atomic sentences therefore partly protect the OOD accuracy.

\paragraph{The $r_2$ target.}
Removing only the $r_2$ target gives the same result as removing all context targets of two-hop sentences (\cref{fig:masks}a).
Adding the $r_2$ target alone to answer-only training lowers the final OOD accuracy in all six seeds (0.37 on average).
Alone it lowers the final OOD accuracy more than together with all other targets (0.37 against 0.58).
With rotary positions removing the $r_2$ target keeps the OOD accuracy in three of four seeds (\cref{app:rotary}).
The $r_2$ target asks the model to predict a relation of the bridge right after its description.
Natural text often continues the mention of an entity in this way.
\Cref{sec:tests} removes the same $r_2$ target from \mbase{}.

\paragraph{Hard targets.}
With hard targets removing the $r_2$ target does not keep the final level of answer-only training (\cref{fig:masks}b).
It raises the peak (0.96 against 0.78) and delays the decline.
Five of six seeds still end below answer-only training.
With hard targets the remaining context targets also lower the final OOD accuracy.

\paragraph{Second hop.}
\Cref{tab:a7-masks} repeats the measurement of \cref{tab:a7} for these conditions.
Where the OOD accuracy is kept, it stays high before the last FFN of loop 2 (0.96--0.98 at epoch 3{,}000).
This FFN then changes almost no correct OOD answer (at most 0.05).
Where it declines, the accuracy before this FFN also falls between epochs 2{,}000 and 3{,}000.
With the $r_2$ target alone the last FFN changes up to 0.75 of the correct OOD answers.

%% file: tables/masks.tex
\begin{tabular}{@{}lccc@{}}
\toprule
targets & peak & final level & below answer-only \\
\midrule
$p_{\mathrm{ref}}$ targets & 0.93\,{\scriptsize(0.83--0.98)} & 0.58\,{\scriptsize(0.21--0.82)} & 6 of 6 \\
without two-hop context targets & 1.00\,{\scriptsize(1.00--1.00)} & 1.00\,{\scriptsize(0.99--1.00)} & 0 of 6 \\
without atomic context targets & 0.95\,{\scriptsize(0.89--0.98)} & 0.41\,{\scriptsize(0.19--0.68)} & 6 of 6 \\
without the $r_2$ target & 1.00\,{\scriptsize(1.00--1.00)} & 0.99\,{\scriptsize(0.99--1.00)} & 0 of 6 \\
answer-only plus the $r_2$ target & 0.94\,{\scriptsize(0.89--0.98)} & 0.37\,{\scriptsize(0.13--0.73)} & 6 of 6 \\
\midrule
hard targets & 0.78\,{\scriptsize(0.50--0.98)} & 0.49\,{\scriptsize(0.14--0.86)} & 6 of 6 \\
hard targets without the $r_2$ target & 0.96\,{\scriptsize(0.87--0.99)} & 0.61\,{\scriptsize(0.46--0.89)} & 5 of 6 \\
answer-only & 0.99\,{\scriptsize(0.98--1.00)} & 0.99\,{\scriptsize(0.96--1.00)} & -- \\
\bottomrule
\end{tabular}

%% file: tables/masks_a7.tex
\begin{tabular}{@{}lccccc@{}}
\toprule
targets & \multicolumn{2}{c}{before the last FFN} & \multicolumn{2}{c}{as trained} & changed to wrong \\
\cmidrule(lr){2-3}\cmidrule(lr){4-5}
 & 2{,}000 & 3{,}000 & 2{,}000 & 3{,}000 & 3{,}000 \\
\midrule
$p_{\mathrm{ref}}$ targets & 0.84 & 0.53 & 0.88 & 0.52 & 0.05--0.59 \\
without two-hop context targets & 0.96 & 0.98 & 1.00 & 0.99 & 0.00--0.01 \\
without atomic context targets & 0.89 & 0.51 & 0.77 & 0.29 & 0.27--0.84 \\
without the $r_2$ target & 0.96 & 0.96 & 1.00 & 1.00 & 0.00--0.00 \\
answer-only plus the $r_2$ target & 0.64 & 0.36 & 0.58 & 0.36 & 0.09--0.75 \\
hard targets & 0.50 & 0.32 & 0.61 & 0.43 & 0.05--0.57 \\
hard targets without the $r_2$ target & 0.88 & 0.45 & 0.91 & 0.54 & 0.02--0.28 \\
answer-only & 0.91 & 0.96 & 0.96 & 0.98 & 0.00--0.05 \\
\bottomrule
\end{tabular}

%% file: sections/appendix/ratio.tex
\section{Twice as many two-hop training sentences}\label[appendix]{app:ratio}

The training set of the main text has 10{,}000 two-hop sentences, two for each of the 5{,}000 atomic facts of \GA{}.
\citet{wang2024grokked} find that this ratio decides how fast a transformer learns to answer held-out two-hop questions.
In their App.~E.2 a model whose lower and upper four layers share their weights slowly learns to answer OOD questions with 12.6 two-hop facts per atomic fact.
\citet{kohli2026loop} observe the same for a looped transformer with about 6.8.
This appendix doubles the two-hop sentences to 20{,}000, four per atomic fact.
It tests whether the baselines of \cref{sec:arch-results} fail only for lack of two-hop sentences.

\paragraph{Training set and runs.}
The training set holds 20{,}000 random chains of \GA{} (\cref{app:data}).
Its residual is 1.24 nats per two-hop sentence (\cref{tab:residual}).
With 10{,}000 atomic sentences an epoch has 59 steps instead of 40.
We train \bloop{} and \btf{} under both losses and \mqal{} under the next-token loss (four seeds each).
We also train \bloop{} and \btf{} under the next-token loss with rotary positions (four seeds each).
All other settings are those of \cref{app:training}.

\begin{table}[htbp]
\caption{Peak, final level and final ID accuracy with 10{,}000 and with 20{,}000 two-hop training sentences (shown as 10{,}000 / 20{,}000).
Means over the same four seeds for both sizes.}
\label{tab:ratio}
\centering
\small
\input{tables/ratio}
\end{table}

\paragraph{Results.}
More two-hop sentences raise the ID accuracy of the baselines but not their OOD accuracy (\cref{tab:ratio}).
Under the answer-only loss \bloop{} now answers the ID questions in all four seeds (final ID accuracy 1.00 against 0.16--0.25).
The final ID accuracy of \btf{} under this loss rises from 0.10 to 0.74.
Under the next-token loss the final ID accuracy rises from 0.03 to 0.15 for \bloop{} and from 0.04 to 0.30 for \btf{}.
With rotary positions it rises from 0.04 to 0.25 for \bloop{} and from 0.05 to 0.30 for \btf{}.
The OOD accuracy of every baseline stays at or below 0.072 at every evaluation.
\bloop{} under the answer-only loss thus fits the ID questions but does not generalize OOD.
Both of our ratios stay below the 6.8 and 12.6 of the studies above.

\mqal{} still rises and then declines.
On average its peak is 0.76 against 0.80 and its final level 0.29 against 0.52.
Per seed the final level is 0.41, 0.08, 0.52 and 0.14 against 0.41, 0.42, 0.39 and 0.86.

%% file: tables/ratio.tex
\begin{tabular}{@{}llllll@{}}
\toprule
positions & model & loss & peak & final level & final ID \\
\midrule
absolute & \bloop{} & answer-only & 0.00 / 0.02 & 0.00 / 0.01 & 0.20 / 1.00 \\
 & \bloop{} & next-token & 0.00 / 0.00 & 0.00 / 0.00 & 0.03 / 0.15 \\
 & \btf{} & answer-only & 0.00 / 0.00 & 0.00 / 0.00 & 0.10 / 0.74 \\
 & \btf{} & next-token & 0.00 / 0.00 & 0.00 / 0.00 & 0.04 / 0.30 \\
 & \mqal{} & next-token & 0.80 / 0.76 & 0.52 / 0.29 & 1.00 / 1.00 \\
\midrule
rotary & \bloop{} & next-token & 0.00 / 0.00 & 0.00 / 0.00 & 0.04 / 0.25 \\
 & \btf{} & next-token & 0.00 / 0.00 & 0.00 / 0.00 & 0.05 / 0.30 \\
\bottomrule
\end{tabular}

%% file: sections/appendix/rotary.tex
\section{Rotary positions}\label[appendix]{app:rotary}

The main text adds a learned position vector to the input of every state.
This appendix repeats its main results with rotary position embeddings (base $10^4$) instead.
No position vector is then added.
Each attention layer rotates the queries and keys of a state by the angle of its position.
$H_i$ and $C_i$ use position $i$.
$Q_i$ uses position $i+1$ in \mq{} and position $i$ in \mqal{} (\cref{sec:design}).
Everything else is as in the main text.

\begin{table}[htbp]
\caption{Peak and final level of the OOD accuracy of the three variants with rotary positions.
Mean over seeds with the range in parentheses.}
\label{tab:rope-variants}
\centering
\small
\input{tables/rotary_variants}
\end{table}

\paragraph{Variants.}
\Cref{tab:rope-variants} repeats \cref{tab:ood} with rotary positions.
The pattern of \cref{sec:arch-results} holds.
Under the next-token loss \mbase{} never generalizes OOD.
\mq{} declines under the answer-only loss (final level 0.07--0.25).
Under the next-token loss \mqal{} ends lower than under the answer-only loss (0.35 against 0.82 on average).
One difference appears.
Under the answer-only loss \mbase{} and \mqal{} also lose part of their OOD accuracy.
\mbase{} ends at least 0.10 below its peak in all four seeds (final levels 0.39--0.83).
\mqal{} does so in four of six seeds (final levels 0.61--0.88).
With absolute positions both keep 0.96--1.00.
Comparing the two losses needs an answer-only reference that keeps its OOD accuracy.
With rotary positions this reference declines in four of six seeds.
The main text therefore uses absolute positions.

\paragraph{Targets.}
\Cref{sec:ladder} reports the four target conditions with both encodings (\cref{fig:ladder,tab:ladder}).
Answer-only training also declines with rotary positions (\cref{tab:rope-variants}).
The comparisons between the conditions are therefore less sharp.
Removing the $r_2$ target from \pref{} targets keeps the OOD accuracy in three of four seeds (final levels 0.99, 0.99 and 0.92).
In the fourth seed it ends at 0.48, below the 0.73 of the same seed with all \pref{} targets.

\begin{table}[htbp]
\caption{As \cref{tab:a7} with rotary positions.
Mean over six seeds and over four seeds in the last row.}
\label{tab:a7-rope}
\centering
\small
\input{tables/rotary_a7}
\end{table}

\paragraph{Second hop.}
\Cref{tab:a7-rope} repeats the measurement of \cref{tab:a7} with rotary positions.
The OOD accuracy before the last FFN of loop 2 again falls with the OOD accuracy as trained.
With \pref{} targets it falls from 0.40 at epoch 2{,}000 to 0.12 at epoch 3{,}000.
The OOD accuracy as trained falls from 0.62 to 0.25 over the same epochs.
With hard targets the two accuracies fall by 0.22 and 0.26.
The last FFN changes up to 0.75 of the correct OOD answers to wrong ones in some seeds.
Without the $r_2$ target the accuracy before this FFN stays at 0.73.
With either encoding most of the decline thus happens before the last FFN.

%% file: tables/rotary_variants.tex
\begin{tabular}{@{}llccc@{}}
\toprule
loss & variant & seeds & peak & final level \\
\midrule
answer-only & \mbase{} & 4 & 0.93\,{\scriptsize(0.85--0.97)} & 0.71\,{\scriptsize(0.39--0.83)} \\
 & \mq{} & 4 & 0.82\,{\scriptsize(0.79--0.87)} & 0.15\,{\scriptsize(0.07--0.25)} \\
 & \mqal{} & 6 & 0.99\,{\scriptsize(0.98--0.99)} & 0.82\,{\scriptsize(0.61--0.96)} \\
\midrule
next-token & \mbase{} & 4 & 0.00\,{\scriptsize(0.00--0.00)} & 0.00\,{\scriptsize(0.00--0.00)} \\
 & \mq{} & 6 & 0.66\,{\scriptsize(0.51--0.90)} & 0.42\,{\scriptsize(0.18--0.60)} \\
 & \mqal{} & 6 & 0.69\,{\scriptsize(0.38--0.96)} & 0.35\,{\scriptsize(0.06--0.93)} \\
\bottomrule
\end{tabular}

%% file: tables/rotary_a7.tex
\begin{tabular}{@{}lccccc@{}}
\toprule
targets & \multicolumn{2}{c}{before the last FFN} & \multicolumn{2}{c}{as trained} & changed to wrong \\
\cmidrule(lr){2-3}\cmidrule(lr){4-5}
 & 2{,}000 & 3{,}000 & 2{,}000 & 3{,}000 & 3{,}000 \\
\midrule
hard targets & 0.47 & 0.25 & 0.58 & 0.32 & 0.03--0.51 \\
$p_{\mathrm{train}}$ targets & 0.59 & 0.49 & 0.72 & 0.61 & 0.01--0.75 \\
$p_{\mathrm{ref}}$ targets & 0.40 & 0.12 & 0.62 & 0.25 & 0.00--0.75 \\
answer-only & 0.95 & 0.85 & 0.97 & 0.89 & 0.00--0.20 \\
$p_{\mathrm{ref}}$ targets without the $r_2$ target & 0.73 & 0.73 & 0.86 & 0.85 & 0.00--0.19 \\
\bottomrule
\end{tabular}

%% file: sections/appendix/seeds.tex
\section{Results of every seed}\label[appendix]{app:seeds}

\Cref{tab:seeds} gives the final level of the OOD accuracy for every seed of the main configurations.
It also gives the selected OOD accuracy of the same runs (\cref{app:training}).
The selected OOD accuracy needs OOD questions for the choice of the snapshot.
We therefore report it only here.
The rows of \mqal{} with hard targets and with answer-only training also serve \cref{sec:ladder}.

\begin{table}[htbp]
\caption{Final level and selected OOD accuracy of every seed (\cref{app:training}).
An empty cell marks a seed that was not run.
A dash marks a run that kept no snapshots.}
\label{tab:seeds}
\centering
\footnotesize
\setlength{\tabcolsep}{3.5pt}
\input{tables/seeds}
\end{table}

%% file: tables/seeds.tex
\begin{tabular}{@{}lcccccc@{\hspace{8pt}}c@{\hspace{8pt}}cccccc@{}}
\toprule
 & \multicolumn{6}{c}{final level} & & \multicolumn{6}{c}{selected OOD accuracy} \\
\cmidrule(lr){2-7}\cmidrule(lr){9-14}
configuration & s0 & s1 & s2 & s3 & s4 & s5 & & s0 & s1 & s2 & s3 & s4 & s5 \\
\midrule
\multicolumn{14}{@{}l}{\emph{\cref{sec:arch-results}}} \\
\quad \mbase{}, answer-only & 0.98 & 0.99 & 1.00 & 0.98 &  &  & & -- & -- & 1.00 & 1.00 &  &  \\
\quad \mbase{}, next-token & 0.00 & 0.00 & 0.00 & 0.00 &  &  & & -- & -- & 0.00 & 0.00 &  &  \\
\quad \mq{}, answer-only & 0.06 & 0.27 & 0.25 & 0.56 &  &  & & 0.88 & 0.85 & 0.90 & 0.98 &  &  \\
\quad \mq{}, next-token & 0.23 & 0.02 & 0.11 & 0.12 &  &  & & 0.81 & 0.72 & 0.62 & 0.90 &  &  \\
\quad \mqal{}, answer-only & 1.00 & 0.98 & 0.99 & 0.99 & 0.96 & 0.99 & & 1.00 & 1.00 & 1.00 & 1.00 & 1.00 & 1.00 \\
\quad \mqal{}, next-token (hard targets) & 0.41 & 0.42 & 0.39 & 0.86 & 0.75 & 0.14 & & 0.92 & 0.55 & 0.94 & 0.93 & 0.99 & 0.53 \\
\quad \bloop{}, answer-only & 0.00 & 0.00 & 0.01 & 0.00 &  &  & & -- & -- & 0.00 & 0.01 &  &  \\
\quad \bloop{}, next-token & 0.00 & 0.00 & 0.00 & 0.00 &  &  & & -- & -- & 0.00 & 0.00 &  &  \\
\quad \btf{}, answer-only & 0.00 & 0.00 & 0.00 & 0.00 &  &  & & 0.00 & 0.00 & 0.00 & 0.00 &  &  \\
\quad \btf{}, next-token & 0.00 & 0.00 & 0.00 & 0.00 &  &  & & 0.00 & 0.00 & 0.00 & 0.00 &  &  \\
\quad \bdisco{}, answer-only & 0.00 & 0.00 & 0.00 & 0.00 &  &  & & 0.00 & 0.00 & 0.00 & 0.00 &  &  \\
\quad \bdisco{}, next-token & 0.00 & 0.00 & 0.00 & 0.00 &  &  & & 0.00 & 0.00 & 0.00 & 0.01 &  &  \\
\midrule
\multicolumn{14}{@{}l}{\emph{\cref{sec:tests}}} \\
\quad \mbase{} without the target of $H_4$ & 0.13 & 0.01 & 0.03 & 0.07 &  &  & & 0.86 & 0.98 & 1.00 & 0.96 &  &  \\
\quad \mq{} with the duplicate slot hidden, answer-only & 0.95 & 0.44 & 0.88 & 0.95 &  &  & & 1.00 & 1.00 & 1.00 & 1.00 &  &  \\
\quad \mq{} with the duplicate slot hidden, next-token & 0.41 & 0.03 & 0.87 & 0.41 &  &  & & 0.97 & 0.60 & 0.96 & 0.99 &  &  \\
\quad \mqal{} with an undecoded slot, answer-only & 0.03 & 0.01 & 0.01 & 0.02 &  &  & & 0.05 & 0.03 & 0.02 & 0.05 &  &  \\
\quad \mqal{} with an undecoded slot, next-token & 0.00 & 0.00 & 0.01 & 0.00 &  &  & & 0.01 & 0.01 & 0.01 & 0.00 &  &  \\
\midrule
\multicolumn{14}{@{}l}{\emph{\cref{sec:ladder}}} \\
\quad \mqal{}, $p_{\mathrm{train}}$ targets & 0.91 & 0.68 & 0.74 & 0.41 & 0.95 & 0.68 & & 0.99 & 0.96 & 0.99 & 0.99 & 0.99 & 0.87 \\
\quad \mqal{}, $p_{\mathrm{ref}}$ targets & 0.73 & 0.63 & 0.82 & 0.70 & 0.21 & 0.37 & & 0.97 & 0.99 & 0.97 & 0.97 & 0.99 & 0.87 \\
\quad \mqal{}, rotary, hard targets & 0.08 & 0.15 & 0.06 & 0.67 & 0.93 & 0.19 & & 0.43 & 0.61 & 0.69 & 0.95 & 0.97 & 0.79 \\
\quad \mqal{}, rotary, $p_{\mathrm{train}}$ targets & 0.21 & 0.79 & 0.08 & 0.97 & 0.94 & 0.78 & & 0.46 & 0.94 & 0.89 & 1.00 & 0.99 & 0.90 \\
\quad \mqal{}, rotary, $p_{\mathrm{ref}}$ targets & 0.15 & 0.73 & 0.17 & 0.11 & 0.18 & 0.54 & & 1.00 & 0.84 & 1.00 & 0.81 & 0.98 & 0.93 \\
\quad \mqal{}, rotary, answer-only & 0.88 & 0.89 & 0.83 & 0.96 & 0.61 & 0.76 & & 1.00 & 1.00 & 1.00 & 1.00 & 1.00 & 1.00 \\
\midrule
\multicolumn{14}{@{}l}{\emph{\cref{app:masks}}} \\
\quad without two-hop context targets & 0.99 & 0.99 & 1.00 & 1.00 & 1.00 & 1.00 & & 1.00 & 1.00 & 1.00 & 1.00 & 1.00 & 1.00 \\
\quad without atomic context targets & 0.36 & 0.21 & 0.68 & 0.61 & 0.38 & 0.19 & & 0.98 & 0.99 & 0.97 & 0.99 & 1.00 & 0.93 \\
\quad without the $r_2$ target & 0.99 & 1.00 & 1.00 & 0.99 & 0.99 & 1.00 & & 1.00 & 1.00 & 1.00 & 1.00 & 1.00 & 1.00 \\
\quad answer-only plus the $r_2$ target & 0.73 & 0.52 & 0.28 & 0.39 & 0.13 & 0.20 & & 0.97 & 0.98 & 0.98 & 0.98 & 0.94 & 0.99 \\
\quad hard targets without the $r_2$ target & 0.51 & 0.48 & 0.79 & 0.89 & 0.57 & 0.46 & & 0.92 & 0.97 & 0.99 & 1.00 & 1.00 & 0.97 \\
\quad rotary, without the $r_2$ target & 0.99 & 0.48 & 0.99 & 0.92 &  &  & & 1.00 & 0.90 & 1.00 & 1.00 &  &  \\
\bottomrule
\end{tabular}